\documentclass[letterpaper]{article} % DO NOT CHANGE THIS

\usepackage{aaai24}

\usepackage{times}  % DO NOT CHANGE THIS
\usepackage{helvet}  % DO NOT CHANGE THIS
\usepackage{courier}  % DO NOT CHANGE THIS
\usepackage[hyphens]{url}  % DO NOT CHANGE THIS
\usepackage{graphicx} % DO NOT CHANGE THIS
\usepackage{natbib}  % DO NOT CHANGE THIS AND DO NOT ADD ANY OPTIONS TO IT
\usepackage{caption} % DO NOT CHANGE THIS AND DO NOT ADD ANY OPTIONS TO IT
\usepackage{algorithm}
\usepackage{algorithmic}

\usepackage{newfloat}
\usepackage{listings}
\DeclareCaptionStyle{ruled}{labelfont=normalfont,labelsep=colon,strut=off} % DO NOT CHANGE THIS
\floatstyle{ruled}
\newfloat{listing}{tb}{lst}{}
\floatname{listing}{Listing}
\usepackage{amsmath}
\usepackage{array}
\usepackage{booktabs}
\usepackage{multirow}
\usepackage{amsfonts}
\usepackage{amssymb}
\usepackage{todonotes}
\usepackage{subfigure}
\usepackage{colortbl}
\usepackage{bm}

\title{Embracing Language Inclusivity and Diversity in CLIP through\\Continual Language Learning}

\author{
    Bang Yang\textsuperscript{\rm 1,2},
    Yong Dai\textsuperscript{\rm 2},
    Xuxin Cheng\textsuperscript{\rm 1},
    Yaowei Li\textsuperscript{\rm 1,2},
    Asif Raza\textsuperscript{\rm 1},
    Yuexian Zou\textsuperscript{\rm 1}\thanks{Corresponding Author.}
}
\affiliations{
    \textsuperscript{\rm 1} ADSPLAB, School of ECE, Peking University, Shenzhen, China\\
    \textsuperscript{\rm 2} Pengcheng Laboratory, Shenzhen, China\\
    \{yangbang, chengxx, ywl, asifraza151, zouyx\}@pku.edu.cn,
    chd-dy@foxmail.com
}

\usepackage{bibentry}
\begin{document}

\maketitle

\begin{abstract}
While vision-language pre-trained models (VL-PTMs) have advanced multimodal research in recent years, their mastery in a few languages like English restricts their applicability in broader communities. To this end, there is an increasing interest in developing multilingual VL models via a joint-learning setup, which, however, could be unrealistic due to expensive costs and data availability. In this work, we propose to extend VL-PTMs' language capacity by continual language learning (CLL), where a model needs to update its linguistic knowledge incrementally without suffering from catastrophic forgetting (CF). We begin our study by introducing a model dubbed CLL-CLIP, which builds upon CLIP, a prevailing VL-PTM that has acquired image-English text alignment. Specifically, CLL-CLIP contains an expandable token embedding layer to handle linguistic differences. It solely trains token embeddings to improve memory stability and is optimized under cross-modal and cross-lingual objectives to learn the alignment between images and multilingual texts. To alleviate CF raised by covariate shift and lexical overlap, we further propose a novel approach that ensures the identical distribution of all token embeddings during initialization and regularizes token embedding learning during training. We construct a CLL benchmark covering 36 languages based on MSCOCO and XM3600 datasets and then evaluate multilingual image-text retrieval performance. Extensive experiments verify the effectiveness of CLL-CLIP and show that our approach can boost CLL-CLIP, e.g., by 6.7\% in text-to-image average Recall@1 on XM3600, and improve various state-of-the-art methods consistently. Our code and data are available at \url{https://github.com/yangbang18/CLFM}.
\end{abstract}

\section{Introduction}
%%%%%%%%%%%%%%%%%%%%%%
\begin{figure}[t]
\centering
\includegraphics[width=0.98\linewidth]{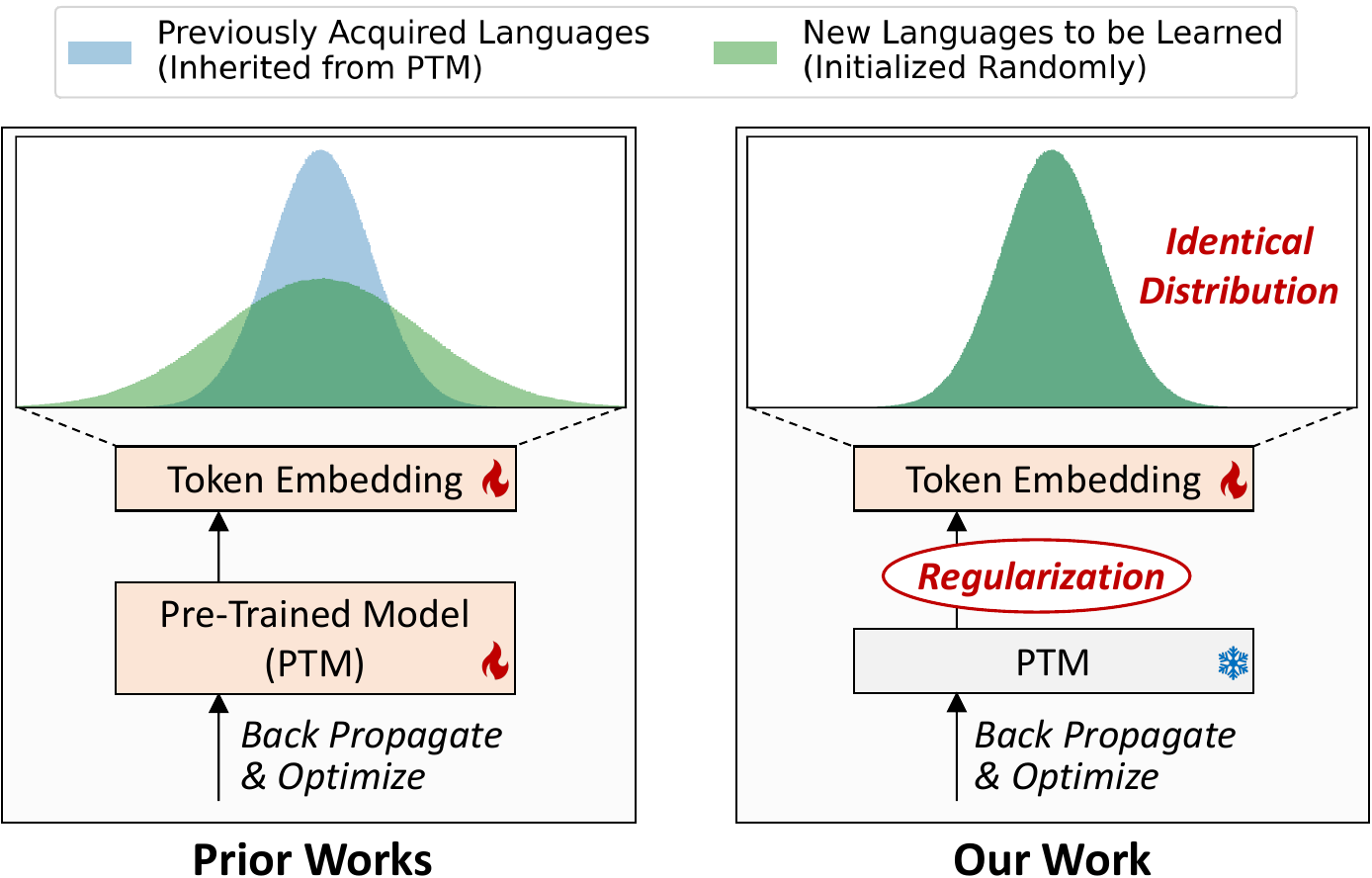}
\caption{For continual language learning, prior works in NLP~\cite{garcia2021continual,huang2022EVS} train full model parameters to learn a new language, with new token embeddings initialized randomly without considering the distribution of prior ones.
Our work requires the least amount of components to be trained (i.e., the token embedding layer) and targets token embedding initialization and regularization to avert catastrophic forgetting. 
Note that our frozen vision PTM is not plotted for clarity.}
\label{fig:intro}
\end{figure}
%%%%%%%%%%%%%%%%%%%%%%

Large-scale vision-language pre-trained models (VL-PTMs) such as CLIP~\cite{radford2021CLIP}, Flamingo~\cite{alayrac2022Flamingo}, and BLIP-2~\cite{li2023BLIP-2} have made great strides in multimodal research~\cite{gan2022VLP,chen2023VLP}. Nevertheless, the majority of the current literature is biased toward a few languages, predominantly English, making it a barrier to the widespread adoption and accessibility of VL-PTMs across different linguistic communities. Considering that we are living in a world with roughly 7,000 languages, it is indispensable to strive for greater language inclusivity and diversity in VL-PTMs.

To endow VL-PTMs with an ability to understand multilingual contexts, there is an increasing interest in developing multilingual VL-PTMs via a joint-learning setup~\cite{zhou2021UC2,zhang2022MLA,chen2023mCLIP,li2023WS-mVLP}, which has shown remarkable performance in tasks like multilingual image-text retrieval. 
However, two critical issues plague the joint learning. 
One is the high computational cost and inflexibility of learning new knowledge, as we need to re-train models on new data alongside all previous data. 
Another one is that data is not always available during the learning cycle due to privacy and other factors. Alternatively, \emph{continual language learning} (CLL), also known as \emph{lifelong language learning}, is a more practical setup to extend PTMs' language capacity with low costs and high flexibility. 
The goal of CLL is to consolidate multilingual performance into a single, parameter- and memory-constrained model, ensuring that this model can evolve under \emph{non-stationary} data streams without suffering from \emph{catastrophic forgetting}~\cite{mccloskey1989CF}. 
While CLL has been extensively studied in natural language processing (NLP)~\cite{biesialska2020LLLSurvey,escolano2021bilingual,zhang2022CLLE,mhamdi2023CCL}, the effective integration of VL-PTMs with CLL is still under-explored and it presents distinctive challenges like leveraging visual information to aid in language learning.

In this paper, we study the multilingual acquisition of VL-PTMs in the CLL setup. We begin our study by selecting CLIP~\cite{radford2021CLIP}, a prevailing VL-PTM that can correlate images and English texts into the same latent space, as our backbone. Next, we propose a model dubbed CLL-CLIP to incrementally learn new languages. Specifically, our model contains an expandable token embedding layer to handle linguistic differences. Such design is crucial to prevent our model from encountering a high portion of \emph{out-of-vocabulary} tokens. 
During training, CLL-CLIP keeps all pre-trained components frozen except its token embedding layer to retain previously acquired knowledge and is optimized under cross-modal and cross-lingual objectives to learn the alignment between images and multilingual texts.

Next, we propose a CLL approach that targets {\bf T}oken {\bf E}mbedding {\bf I}nitialization and {\bf R}egularization (TEIR) to alleviate catastrophic forgetting (CF). 
Figure~\ref{fig:intro} differentiates our TEIR from prior approaches in NLP~\cite{garcia2021continual,huang2022EVS}. 
In particular, to reduce CF raised by \emph{covariate shift}~\cite{shimodaira2000CovariateShift,ioffe2015BN}, our approach ensures the \emph{identical distribution} of all token embeddings during initialization. 
To mitigate CF caused by the \emph{lexical overlap}~\cite{pfeiffer2021UNKsEverywhere}, our approach regularizes token embedding learning based on the number of times that tokens appear in the tasks they have already learned by CLL-CLIP. 
Our insight is that if a token is common in previously learned tasks, its embedding update should be penalized to avoid task interference.

To evaluate the effectiveness of our CLL-CLIP model and TEIR approach, we first construct a benchmark covering 36 languages based on MSCOCO~\cite{chen2015MSCOCO} and XM3600~\cite{thapliyal2022XM3600} datasets. 
We then re-produce various state-of-the-art (SOTA) continual learning and parameter-efficient fine-tuning methods based on our CLL-CLIP model on this benchmark. Extensive experiments verify the effectiveness of CLL-CLIP and show that TEIR can boost CLL-CLIP, e.g., by 6.7\% in text-to-image average Recall@1 on XM3600, and improve the performance of SOTA methods consistently.

Our main contributions are as follows. (1) To the best of our knowledge, we present the first systematic study on enhancing the language capacity of dual-stream VL-PTMs through continual language learning. (2) We design a model named CLL-CLIP for this challenging setup and introduce a novel approach called TEIR that underscores the initialization and regularization of token embeddings to mitigate catastrophic forgetting. (3) We construct a CLL benchmark for evaluating image-text retrieval across 36 languages. Extensive experiments verify the effectiveness of our CLL-CLIP and TEIR and demonstrate the generality of TEIR on various SOTA methods.

\section{Related Work}

\paragraph{Multilingual VL Pre-Training} 
As monolingual visual-language pre-training models (VL-PTMs) continue to evolve, an increasing amount of effort is directed toward enhancing the adaptability of these models for multilingual scenarios via pre-training. 
M$^3$P~\cite{ni2021M3P} and UC$^2$~\cite{zhou2021UC2} adopt a BERT-like single-stream architecture~\cite{devlin2019BERT} for pre-training, yet they diverge in their data augmentation strategies. 
M$^3$P uses word-level augmentation to obtain code-switched VL pairs, whereas UC$^2$ utilizes translation engines to transform English image captions into other languages. 
In contrast, MURAL~\cite{jain2021MURAL}, M-CLIP~\cite{carlsson2022MCLIP}, MLA~\cite{zhang2022MLA}, and mCLIP~\cite{chen2023mCLIP} build their model on a dual-stream model like CLIP for better efficiency on retrieval tasks. 
These models use the same data augmentation strategy as UC$^2$, but MURAL and mCLIP additionally consider annotated translation pairs. 
Besides retrieval tasks, recent encoder-decoder-based PaLI~\cite{chen2023PaLI} and WS-mVLP~\cite{li2023WS-mVLP} have shown their superiority in multilingual VL generation tasks. 
However, all the above methods develop multilingual VL-PTMs via a joint-learning setup and thus suffer from high costs and inflexibility of learning new languages. 
In this paper, we focus on endowing dual-stream VL-PTMs with a multilingual understanding ability via a more practical and flexible setup, i.e., continual language learning.

%%%%%%%%%%%%%%%%%%%%%
\begin{figure*}[t]
\centering
\includegraphics[width=0.9\linewidth]{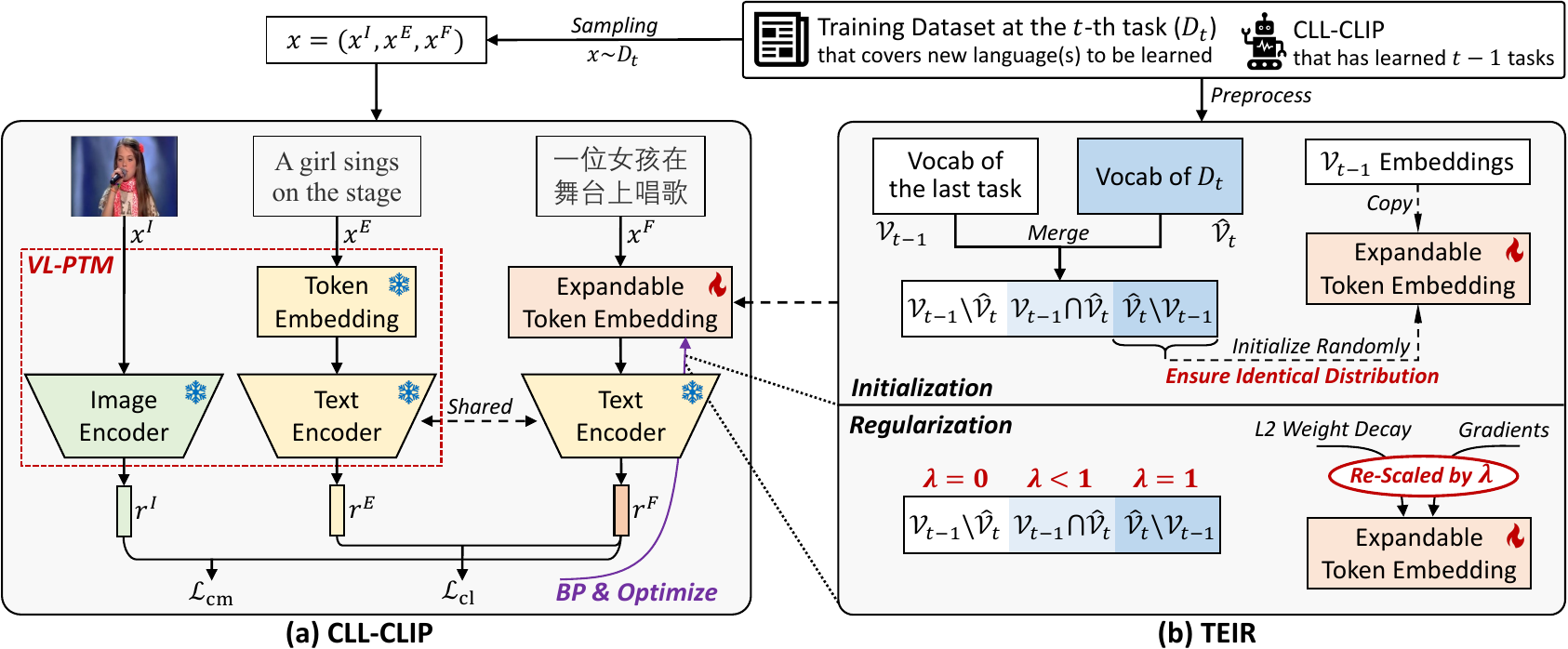}
\caption{
Overview of our proposals. (a): CLL-CLIP builds upon a two-tower VL-PTM (i.e., CLIP), keeps all pre-trained components frozen, and contains an expandable and trainable token embedding layer for continual language learning. (b): Our TEIR approach eases catastrophic forgetting by underscoring the initialization and regularization of token embeddings.
}
\label{fig:framework}
\end{figure*}
%%%%%%%%%%%%%%%%%%%%%

\paragraph{Continual Learning (CL)} The core aspiration of CL is to enable machines to mimic the strong adaptability of humans to continually acquire, update, organize, and exploit knowledge~\cite{wang2023CLSurvey}. The computer vision (CV) community has witnessed significant advances in CL, which can be mainly divided into four categories. Specifically, 
\emph{regularization}-based methods penalize changes to model parameters or predictions~\cite{kirkpatrick2017EWC,lee2019GD,ahn2021SS-IL}; 
\emph{rehearsal}-based methods store historical data or features to retain previously acquired knowledge~\cite{chaudhry2019ER,buzzega2020DER,cha2021Co2L}; 
\emph{architecture}-based methods assign isolated parameters for different tasks~\cite{yoon2018DEN,li2019Learn2Grow,ke2020CAT}; 
\emph{prompt}-based methods add parameter-efficient modules into frozen PTMs to harness their power~\cite{wang2022DualPrompt,wang2022L2P,smith2023CodaPrompt,gao2023LAE}. 
The success of CL in CV inspires related research in NLP~\cite{biesialska2020LLLSurvey,wu2022PLMinCL,mhamdi2023CCL}. In particular, a line of research studies on how to add new languages to pre-trained neural machine translation models. One attempt is to add and train language-specific components, like encoder/decoder~\cite{escolano2021bilingual} and adapter~\cite{berard2021continual}. Another attempt proposes to substitute models' vocabulary dynamically~\cite{garcia2021continual,huang2022EVS}. 
In this paper, we differentiate our work from prior ones in NLP in Figure~\ref{fig:intro}. 
Unlike those regularization methods that need to estimate parameter importance by feeding data into the model, our approach only requires the lexical statistics of data. 
By contrast with the CL of CLIP in visual recognition~\cite{ding2022DontStopLearning,thengane2022clip}, we value the CL of CLIP in language acquisition.

\section{Approach}
In our continual language learning (CLL) setting, a model needs to sequentially learn $T$ tasks, each with its corresponding training dataset $D_t (t \in [1, T])$ that covers non-overlapping subsets of languages. After training a model parameterized by $\bm{\phi}_t$ on $D_t$, the goal of CLL is to ensure the model can perform well in previous $t$ tasks. To achieve that, we propose CLL-CLIP and TEIR, as introduced next.

\subsection{CLL-CLIP}
\subsubsection{Architecture}
As shown in Figure~\ref{fig:framework}(a), our model builds upon CLIP to avoid from-scratch training and contains an expandable token embedding layer parameterized by $\bm{\theta}_t$ to vectorize multilingual texts. In particular, CLIP consists of a vision encoder, a text encoder, and a token embedding layer mainly for English\footnote{We separate token embeddings from CLIP for clarity, and the text encoder means the rest of the components (positional embeddings, Transformer blocks, projection head) to obtain text features.}. Let denote their parameters as $\bm{\Omega}_{ve}$, $\bm{\Omega}_{te}$, and $\bm{\Omega}_{emb}$, respectively. Then parameters of our model at the $t$-th task are $\bm{\phi}_t =\{\bm{\Omega}_{ve}, \bm{\Omega}_{te}, \bm{\Omega}_{emb}, \bm{\theta}_t\}$, where $\bm{\Omega}_{emb}$ can be discarded during inference. 
We keep all CLIP parameters $\bm{\Omega}_*$ frozen and solely train $\bm{\theta}_t$. This choice is in line with the research on efficient VL pre-training~\cite{zhai2022LiT,zhang2022MLA} and also benefits the preservation of previously acquired knowledge during continual learning~\cite{wang2022L2P,smith2023CodaPrompt}.

\subsubsection{Vocab Substitution} Let denote the vocab corresponding to $\bm{\theta}_t$ as $\mathcal{V}_t$.
Before training, $\mathcal{V}_0$ is identical to CLIP's vocab, and $\bm{\theta}_0 = \bm{\Omega}_{emb}$. For $t \in [1, T]$, $\mathcal{V}_t$ needs to be dynamically updated to accommodate the lexicon of new languages. 
Thus, we first adopt the same BPE procedure~\cite{sennrich2016BPE} as CLIP to build vocab $\hat{\mathcal{V}}_t$ from $D_t$ and then follow \cite{garcia2021continual} to obtain $\mathcal{V}_t$ by merging $\mathcal{V}_{t-1}$ and $\hat{\mathcal{V}}_t$, i.e., $\mathcal{V}_t = \mathcal{V}_{t-1} \cup \hat{\mathcal{V}}_t$. 
There are two issues to be noted: (1) the embedding initialization of $\hat{\mathcal{V}}_t \setminus \mathcal{V}_{t-1}$ (new tokens that only exist in $\hat{\mathcal{V}}_t$) and (2) the sub-optimal nature of $\mathcal{V}_t$ due to lacking comprehensive text statistics. 
We will address (1) in TEIR and discuss (2) in later experiments.

\subsubsection{Training Objectives}
Each training sample for our CLL-CLIP is a triplet $x = (x^I, x^E, x^F)$ that includes an image $x^I$, a text in \emph{native} language $x^E$ (i.e., \emph{English} text), and a \emph{foreign} text $x^F$. 
At the $t$-th task, we obtain global representations of the triplet $x$ as follows:
\begin{equation}
\begin{aligned} 
   \bm{r}^I &= g(x^I; \bm{\Omega}_{ve}),\\
   \bm{r}^E &= g(x^E; \bm{\Omega}_{te}, \bm{\Omega}_{emb}),\\
   \bm{r}^F &= g(x^F; \bm{\Omega}_{te}, \bm{\theta}_t),
\end{aligned}
\end{equation}
where $g(\cdot)$ indicates the feed-forward transformation. We suggest training CLL-CLIP with cross-modal and cross-lingual objectives, i.e., $\mathcal{L}_{\text{cm}}$ and $\mathcal{L}_{\text{cl}}$, so that CLL-CLIP can correlate $\bm{r}^I$ with $\bm{r}^F$ based on the already acquired knowledge, i.e., the alignment between $\bm{r}^I$ and $\bm{r}^E$. Following CLIP, we implement $\mathcal{L}_{\text{cm}}$ as InfoNCE-based image-text contrast~\cite{oord2018infoNCE}:
\begin{equation}
\begin{aligned} 
\mathcal{L}_{\text{cm}} &= \frac{1}{2} \left(
\mathcal{L}_{\text{InfoNCE}}^{I \rightarrow F} + \mathcal{L}_{\text{InfoNCE}}^{F \rightarrow I}
\right), \\
\mathcal{L}_{\text{InfoNCE}}^{Y \rightarrow Z} &= 
- \frac{1}{K} \sum_{k=1}^K \log{\frac{\exp(\langle \bm{r}^Y_k, \bm{r}^Z_k\rangle / \tau)}{\sum_{l=1}^{K} \exp(\langle \bm{r}^Y_k, \bm{r}^Z_l\rangle / \tau)}},
\end{aligned}
\end{equation}
where $K$ denotes the batch size, $\langle \cdot, \cdot \rangle$ the cosine similarity, and $\tau$ a temperature hyper-parameter.
Motivated by \cite{reimers2020MKD}, we implement $\mathcal{L}_{\text{cl}}$ as the mean-square error between paired text features:
\begin{equation}
\label{eq:cross_lingual}
\mathcal{L}_{\text{cl}} = \frac{1}{2K} \sum_{k=1}^K ||\bm{r}^E_k - \bm{r}^F_k||_2^2,
\end{equation}
where $||\cdot||_2$ denotes L2-norm. The overall training objective of CLL-CLIP can be formulated as follows:
\begin{equation}
\label{eq:overall_loss}
\mathcal{L} = \gamma_1 \cdot \mathcal{L}_{\text{cm}} + \gamma_2 \cdot \mathcal{L}_{\text{cl}},
\end{equation}
where $\gamma_*$ are hyper-parameters to balance two losses.

\subsection{TEIR}
As shown in Figure~\ref{fig:framework}(b), the key of TEIR is how we treat $\mathcal{V}_{t, old} = \mathcal{V}_{t-1} \setminus \hat{\mathcal{V}}_t$, $\mathcal{V}_{t,\cap} = \mathcal{V}_{t-1} \cap \hat{\mathcal{V}}_t$, and $\mathcal{V}_{t, new} = \hat{\mathcal{V}}_t \setminus \mathcal{V}_{t-1}$ differently to mitigate catastrophic forgetting (CF). 

\subsubsection{Initialization} 
Language models building on Transformer \cite{vaswani2017Transformer} typically initialize token embeddings with a Gaussian distribution $\mathcal{N}(\mu, \sigma^2)$ with zero mean ($\mu=0$) and a pre-defined variance $\sigma^2$. 
Let denote CLL-CLIP's token embeddings after training on $D_t$ as $\bm{\theta}^*_t$. With the assumption that $\bm{\theta}^*_{t-1} \sim \mathcal{N}(\mu_{t-1}, \sigma_{t-1}^2)$, the focus now becomes how we initialize $\bm{\theta}_t$ properly. Following~\cite{garcia2021continual}, $\bm{\theta}_t$ inherits pre-trained embeddings of $\mathcal{V}_{t-1}$ from $\bm{\theta}^*_{t-1}$ to preserve previously acquired linguistic knowledge. Instead of initializing embeddings of $\mathcal{V}_{t,new}$ with a fixed distribution $\mathcal{N}(\mu, \sigma^2)$, we suggest $\mu = \mu_{t-1}$ and $\sigma = \sigma_{t-1}$ to ensure the \emph{identical distribution} of new and prior token embeddings. By doing so, our approach alleviates the feature drift (a.k.a. covariate shift) problem, which is a potential factor to arise CF~\cite{ramasesh2021AnatomyCF}.

\subsubsection{Regularization} Although \emph{lexical overlap} is beneficial for transfer learning~\cite{pfeiffer2021UNKsEverywhere}, learning the embeddings of $\mathcal{V}_{t, \cap}$ without constraints will cause interference to the performance of previous tasks that contain lexically overlapping tokens. Let denote token statistics till the $t$-th task as $\bm{c}_t \in \mathbb{R}^{|\mathcal{V}_t|}$, where $c_{t,j}$ is the number of times that the $j$-th token appears in prior $t-1$ tasks and $c_{1, j}$ is initialized as 1. To overcome CF raised by lexical overlap, we re-scale the rate of L2 weight decay $\beta$ and gradients $\nabla\mathcal{L}(\bm{\theta}_t)$ w.r.t. token embeddings $\bm{\theta}_t$ as follows, with standard stochastic gradient descent (SGD) with L2 weight decay as an example:
\begin{equation}
\label{eq:sgd}
\bm{\theta}_{t, j} \leftarrow (1 - \alpha\beta\lambda_{t, j})\bm{\theta}_{t, j} - \alpha\lambda_{t, j}\nabla\mathcal{L}(\bm{\theta}_{t, j})
\end{equation}
where $\alpha$ is a learning rate, and $\lambda_{t, j}$ is defined as:
\begin{equation}
\label{eq:coeff}
\lambda_{t, j} =
\begin{cases}
0,  & \text{if token$_j$} \in \mathcal{V}_{t, old}\\
1/(c_{t,j} + 1), & \text{if token$_j$} \in \mathcal{V}_{t, \cap}\\
1, & \text{if token$_j$} \in \mathcal{V}_{t, new}\\
\end{cases}
\end{equation}
For sophisticated optimizers with momentum, the scaling operation is still applied on $\beta$ and $\nabla\mathcal{L}(\bm{\theta}_t)$ directly. As indicated by Equation~(\ref{eq:sgd}) and (\ref{eq:coeff}), we keep token embeddings unrelated to the $t$-th task intact, penalize embedding learning of $\mathcal{V}_{t, \cap}$, while updating embeddings of $\mathcal{V}_{t, new}$ as usual. This method averts task interference and ensures the effective learning of text features ($\bm{r}^F$), leading to a better trade-off between memory stability and learning plasticity.

%%%%%%%%%%
\begin{table}[t]
    \centering
    % \fontsize{8}{10}\selectfont
    \small
    \setlength\tabcolsep{4pt}
    \begin{tabular}{lccccc}  
    \toprule
    & MSCOCO$_{36}$ & XM3600\cr
    \midrule
    \# Train/Val/Test Images       &113,287/5,000/5,000    &-/-/3,600\cr
    \# Languages                &1 + 35                 &36\cr
    \# Captions per Language    &616,767                &$\approx$7260\cr
    \bottomrule
    \end{tabular}
    \caption{Dataset Statistics. \# means ``The number of". MSCOCO$_{36}$ is obtained by translating the English captions of MSCOCO into the other 35 languages in XM3600 via Google Translator, following \cite{thapliyal2022XM3600}.}
    \label{tab:statistics} 
\end{table}
%%%%%%%%%%

%%%%%%%%%%
\begin{table*}[t]
\centering
\setlength{\tabcolsep}{3.5pt}
% \fontsize{8}{9}\selectfont
% \small
\begin{tabular}{@{}ll llll llll@{}}
\toprule 
\multirow{3}{*}[-5pt]{\begin{tabular}[c]{@{}c@{}}Setting\end{tabular}} 
&\multirow{3}{*}[-5pt]{Model} 
&\multicolumn{4}{c}{MSCOCO$_{36}$ (In-Domain)}
&\multicolumn{4}{c}{XM3600 (Out-of-Domain)}
\\
\cmidrule(lr){3-6} 
\cmidrule(lr){7-10}
&
&\multicolumn{2}{c}{Image-to-Text}
&\multicolumn{2}{c}{Text-to-Image}
&\multicolumn{2}{c}{Image-to-Text}
&\multicolumn{2}{c}{Text-to-Image}
\\ 
\cmidrule(lr){3-4} 
\cmidrule(lr){5-6}
\cmidrule(lr){7-8} 
\cmidrule(lr){9-10}
&
&AR ($\uparrow$) &F ($\downarrow$) 
&AR ($\uparrow$) &F ($\downarrow$)
&AR ($\uparrow$) &F ($\downarrow$) 
&AR ($\uparrow$) &F ($\downarrow$)
\\
\midrule% [\heavyrulewidth] 

\multirow{3}{*}{\begin{tabular}[c]{@{}l@{}}Joint\\Learning\end{tabular}}

&CLL-CLIP
&\bf 53.3 &-
&\bf 31.4 &-
&50.7 &-
&37.1 &-
\\

&M-CLIP~\citeyearpar{carlsson2022MCLIP}
&42.7 &-
&25.9 &-
&\bf 53.6 &-
&\bf 41.1 &-
\\

&PaLI~\citeyearpar{chen2023PaLI}
&- &-
&- &-
&36.0 &-
&28.5 &-
\\
\midrule

\multirow{18}{*}{\begin{tabular}[c]{@{}l@{}}Continual\\Learning\end{tabular}}
&CLL-CLIP
&29.6 &23.2
&15.2 &15.6
&26.4 &23.1
&17.6 &18.4
\\

&\quad \bf with TEIR
&\bf 38.3 \scriptsize (+8.7) 
&\bf 14.7 \scriptsize (+8.5)
&\bf 20.5 \scriptsize (+5.3)
&\bf 10.5 \scriptsize (+5.1)
&\bf 35.0 \scriptsize (+8.6) 
&\bf 15.3 \scriptsize (+7.8)
&\bf 24.3 \scriptsize (+6.7)
&\bf 12.5 \scriptsize (+5.9)

\\
\cmidrule(lr){2-10}

&oEWC \citeyearpar{schwarz2018oEWC}
&37.0 &15.7
&19.3 &11.3
&32.3 &17.2
&21.8 &14.1
\\
&\quad \bf with TEIR
&\bf 40.2 \scriptsize (+3.2)
&\bf 12.7 \scriptsize (+3.0)
&\bf 21.6 \scriptsize (+2.3)
&\bf 9.3 \scriptsize (+2.0)
&\bf 36.7 \scriptsize (+4.4)
&\bf 13.4 \scriptsize (+3.8)
&\bf 25.6 \scriptsize (+3.8)
&\bf 11.2 \scriptsize (+2.9)
\\
\cmidrule(lr){2-10}

&ER \citeyearpar{chaudhry2019ER}
&34.1 &17.9
&17.8 &12.3
&29.0 &20.0
&19.4 &16.0
\\
&\quad \bf with TEIR
&\bf 39.3 \scriptsize (+5.2) 
&\bf 12.8 \scriptsize (+5.1)
&\bf 21.5 \scriptsize (+3.7)
&\bf 8.8 \scriptsize (+3.5)
&\bf 35.4 \scriptsize (+6.4) 
&\bf 13.9 \scriptsize (+6.1)
&\bf 24.7 \scriptsize (+5.3)
&\bf 11.2 \scriptsize (+4.8)
\\
\cmidrule(lr){2-10}

&DER \citeyearpar{buzzega2020DER}
&37.6 &14.6
&19.5 &10.6
&31.6 &17.4
&21.0 &14.4
\\
&\quad \bf with TEIR
&\bf 42.7 \scriptsize (+5.1) 
&\bf 9.4 \scriptsize (+5.2)
&\bf 23.4 \scriptsize (+3.9) 
&\bf 6.9 \scriptsize (+3.7)
&\bf 38.3 \scriptsize (+6.7) 
&\bf 10.9 \scriptsize (+6.5)
&\bf 26.7 \scriptsize (+5.7) 
&\bf 9.3 \scriptsize (+5.1)
\\
\cmidrule(lr){2-10}

&MLA$\dagger$ \citeyearpar{zhang2022MLA}
&35.9 &20.9
&18.4 &15.0
&30.7 &21.8
&20.6 &18.1
\\

&\quad \bf with TEIR
&\bf 46.0 \scriptsize (+10.1)
&\bf 11.2 \scriptsize (+9.7)
&\bf 25.2 \scriptsize (+6.8)
&\bf 8.6 \scriptsize (+6.4)
&\bf 41.1 \scriptsize (+10.4)
&\bf 12.3 \scriptsize (+9.5)
&\bf 29.0 \scriptsize (+8.4)
&\bf 10.7 \scriptsize (+7.4)
\\
\cmidrule(lr){2-10}

&P-Tuning$\dagger$ \citeyearpar{liu2022P-Tuning}
&30.1 &23.9
&15.0 &16.3
&24.9 &23.9
&16.4 &19.3
\\
&\quad \bf with TEIR
&\bf 41.1 \scriptsize (+11.0)
&\bf 13.3 \scriptsize (+10.6)
&\bf 22.2 \scriptsize (+7.2)
&\bf 9.6 \scriptsize (+6.7)
&\bf 35.5 \scriptsize (+10.6)
&\bf 13.8 \scriptsize (+10.1)
&\bf 25.4 \scriptsize (+9.0)
&\bf 11.5 \scriptsize (+7.8)
\\
\cmidrule(lr){2-10}

&LoRA$\dagger$ \citeyearpar{hu2022LoRA}
&31.8 &22.5
&16.2 &15.9
&28.0 &22.7
&18.7 &18.9

\\
&\quad \bf with TEIR
&\bf 41.6 \scriptsize (+9.8)
&\bf 12.9 \scriptsize (+9.6)
&\bf 22.8 \scriptsize (+6.6)
&\bf 9.7 \scriptsize (+6.2)
&\bf 38.0 \scriptsize (+10.0)
&\bf 13.9 \scriptsize (+8.8)
&\bf 27.0 \scriptsize (+8.3)
&\bf 11.7 \scriptsize (+7.2)
\\
\cmidrule(lr){2-10}

&DualPrompt \citeyearpar{wang2022DualPrompt}  
&28.4 &23.6
&14.1 &15.8
&25.5 &22.9
&16.4 &18.4
\\
&\quad \bf with TEIR
&\bf 38.3 \scriptsize (+9.9)
&\bf 14.0 \scriptsize (+9.6)
&\bf 19.7 \scriptsize (+5.6)
&\bf 10.6 \scriptsize (+5.2)
&\bf 35.3 \scriptsize (+9.8)
&\bf 14.1 \scriptsize (+8.8)
&\bf 23.6 \scriptsize (+7.2)
&\bf 12.1 \scriptsize (+6.3)
\\
\cmidrule(lr){2-10}

&CodaPrompt \citeyearpar{smith2023CodaPrompt}  
&28.9 &22.6
&14.4 &15.2
&24.6 &22.2
&15.9 &17.6
\\
&\quad \bf with TEIR
&\bf 41.4 \scriptsize (+12.5)
&\bf 9.7 \scriptsize (+12.9)
&\bf 22.3 \scriptsize (+7.9)
&\bf 7.1 \scriptsize (+8.1)
&\bf 36.7 \scriptsize (+12.1)
&\bf 9.3 \scriptsize (+12.9)
&\bf 25.3 \scriptsize (+9.4)
&\bf 7.9 \scriptsize (+9.7)
\\

\bottomrule
\end{tabular}

\caption{Retrieval performance on MSCOCO$_{36}$ and XM3600. $\dagger$: Task identity is needed during inference. All results are re-produced by ourselves except that of PaLI. Note that PaLI is not optimized for image-text retrieval, but we draw its results from \cite{chen2023PaLI} for completeness. The numbers in brackets indicate the absolute improvements brought by our approach.}
\label{tab:main}

\end{table*}
%%%%%%%%%%

\section{Experiments}
\subsection{Experimental Settings}
\subsubsection{Benchmark}
We build a CLL benchmark based on \textbf{MSCOCO} \cite{chen2015MSCOCO} and \textbf{XM3600} \cite{thapliyal2022XM3600} to evaluate the effectiveness of our proposals. Here are the reasons: (1) MSCOCO is a popular VL benchmark and it contains high-quality image-English caption pairs. (2) XM3600 consists of image-caption pairs in 36 languages\footnote{The 36 languages are Arabic, Bengali, Czech, Danish, German, Greek, English, Spanish, Farsi, Finnish, Filipino, French, Hebrew, Hindi, Croatian, Hungarian, Indonesian, Italian, Japanese, Korean, Maori, Dutch, Norwegian, Polish, Portuguese, Cusco Quechua, Romanian, Russian, Swedish, Swahili, Telugu, Thai, Turkish, Ukrainian, Vietnamese, and Chinese-Simplified.} spoken by geographically-diverse people. This dataset covers the most diverse languages to our best knowledge. (3) The multi-lingual VL benchmark IGLUE~\cite{bugliarello2022IGLUE} varies in both task types and languages, making it hard to justify the effect of linguistic differences.
As shown in Table~\ref{tab:statistics}, we use Google Translator\footnote{https://translate.google.com/} for data augmentation and thus obtain a multilingual dataset named MSCOCO$_{36}$. We train models on MSCOCO$_{36}$ based on the Karpathy split~\cite{karpathy2015deep}. Then, we report \emph{in-domain} and \emph{out-of-domain} results on MSCOCO$_{36}$ and XM3600, respetively.

\subsubsection{Tasks and Task Order} We treat each language as a task and thus obtain $T=36$ tasks. Models are trained on the English task first and then the rest 35 tasks in a random order. 

\subsubsection{Metrics} Let $a_{j,i}$ ($j \geq i$) denotes Recall@1 (a popular metric in information retrieval) on the $i$-th task after training on the $j$-th task. In line with the continual learning research \cite{wang2023CLSurvey}, we compute two metrics: 
\begin{itemize}
\item Average Recall: $\mathbf{AR}_j = \frac{1}{j}\sum_{i=1}^{j}a_{j,i}$, a composite metric for a model's learning capacity and memory stability.
\item Forgetting: $\mathbf{F}_j = \frac{1}{j-1}\sum_{i=1}^{j-1}\mathop{\max}_{k\in[1, j-1]}(a_{k,i} - a_{j, i})$, whose lower value means less catastrophic forgetting.
\end{itemize}
Unless otherwise specified, we report the-end $\mathbf{AR}_T$ and $\mathbf{F}_T$ performance in \emph{percentile} and omit the subscript.

\subsubsection{Implementation Details} We follow \cite{zhang2022MLA,yang2023MultiCapCLIP} to adopt the ViT-B/16 variant of CLIP as the backbone. 
We follow OpenCLIP~\cite{ilharco2021openclip} and set the initial temperature of $\mathcal{L}_{\text{cm}}$ to 0.07. We search the hyperparameters $\gamma_1$ and $\gamma_2$ in Equation (\ref{eq:overall_loss}) from values $\{1, 0.1, 0.01\}$ and set $\gamma_1 = 0.01$ and $\gamma_2 = 1$ based on the AR metric on the validation set. For models without TEIR, we initialize new token embeddings with $\mathcal{N}(0, 0.02^2)$ following OpenCLIP and set $\forall t, \forall j, \lambda_{t,j} = 1$ (Equation (\ref{eq:coeff})).
For each task, we set the vocab size to 10K. We use batches of 128 samples and AdamW \cite{loshchilov2019AdamW} with L2 weight decay of 0.05 to train models for 3 epochs. We set the learning rate fixed to 5e-5 after 10\% warm-up iterations. The model achieving the highest summation of Recall@\{1, 5, 10\} on the current-task validation set is selected for training on the next task. We conduct experiments in PyTorch on a single NVIDIA V100 card and every run of an experiment takes less than 20 hours.

\subsubsection{Comparing Methods} We reproduce the following SOTA continual learning (CL) and parameter-efficient fine-tuning (PEFT) methods for comparisons: (1) regularization-based online Elastic Weight Consolidation (\textbf{oEWC})~\cite{schwarz2018oEWC} that penalizes the changes in model parameters; (2) rehearsal-based \textbf{ER}~\cite{chaudhry2019ER} that stores historical training samples for current-task learning; (3) rehearsal- and regularization-based \textbf{DER}~\cite{buzzega2020DER} that stores features of previously learned samples for knowledge distillation; (4) architecture-based \textbf{MLA}~\cite{zhang2022MLA}, \textbf{P-Tuning}~\cite{liu2022P-Tuning}, and \textbf{LoRA}~\cite{hu2022LoRA} that inserts task-specific adapters~\cite{houlsby2019Adapter}, learnable prompt tokens, and decomposed matrices into frozen PTMs, respectively. (5) prompt-based \textbf{DualPrompt}~\cite{wang2022DualPrompt} and \textbf{CodaPrompt}~\cite{smith2023CodaPrompt} that rely on a key-query mechanism to generate proper prompts for frozen PTMs. We reproduce all the above methods in the text branch of CLL-CLIP with the aforementioned implementation details.

%%%%%%%%%%%%%%
\begin{table*}[t]
\centering
% \small
\setlength{\tabcolsep}{4pt}
% \fontsize{8}{9}\selectfont
\begin{tabular}{@{}lccc c cc cc cc cc@{}}
\toprule 

\multirow{3}{*}[-5pt]{\begin{tabular}[c]{@{}c@{}}Setting \end{tabular}}
&Initialization
&\multicolumn{2}{c}{Regularization}
&\multirow{3}{*}[-5pt]{\begin{tabular}[c]{@{}c@{}}Oracle\\Vocab \end{tabular}}
&\multicolumn{4}{c}{MSCOCO$_{36}$ (In-Domain)}
&\multicolumn{4}{c}{XM3600 (Out-of-Domain)}
\\
\cmidrule(lr){2-2}
\cmidrule(lr){3-4}
\cmidrule(lr){6-9} 
\cmidrule(lr){10-13} 
&\multirow{2}{*}[-3pt]{\begin{tabular}[c]{@{}c@{}}Identical\\Distribution \end{tabular}}
&\multirow{2}{*}[-3pt]{Gradient}
&\multirow{2}{*}[-3pt]{\begin{tabular}[c]{@{}c@{}}Weight\\Decay \end{tabular}}
&
&\multicolumn{2}{c}{Image-to-Text}
&\multicolumn{2}{c}{Text-to-Image}
&\multicolumn{2}{c}{Image-to-Text}
&\multicolumn{2}{c}{Text-to-Image}
\\ 
\cmidrule(lr){6-7} 
\cmidrule(lr){8-9}
\cmidrule(lr){10-11} 
\cmidrule(lr){12-13}
&&&&
&AR ($\uparrow$) &F ($\downarrow$)
&AR ($\uparrow$) &F ($\downarrow$)
&AR ($\uparrow$) &F ($\downarrow$)
&AR ($\uparrow$) &F ($\downarrow$)
\\

\midrule% [\heavyrulewidth]

(1): CLL-CLIP & & &
&$\times$
&29.6 &23.2
&15.2 &15.6
&26.4 &23.1
&17.6 &18.4

\\

(2) &$\surd$ & &
&$\times$
&32.4 &21.9
&16.8 &15.1
&29.9 &22.5
&20.5 &18.2
\\

(3) & &$\surd$ &
&$\times$
&31.9 &20.3
&16.9 &13.5
&28.4 &20.2
&19.5 &16.1
\\

(4) & & &$\surd$
&$\times$
&33.3 &19.2
&17.0 &13.5
&30.0 &19.0
&19.9 &15.5
\\

(5) &&$\surd$ &$\surd$ 
&$\times$
&37.2 &14.9
&19.7 &10.6
&33.3 &15.4
&22.8 &12.6
\\

% \rowcolor{gray!10}
(6): (1) + TEIR &$\surd$ 
&$\surd$ &$\surd$ 
&$\times$
&\bf 38.3 &\bf 14.7
&\bf 20.5 &\bf 10.5
&\bf 35.0 &\bf 15.3
&\bf 24.3 &\bf 12.5
\\

\midrule
(7) &$\surd$ &$\surd$ &$\surd$ 
&$\surd$
&\bf 42.4 &\bf 10.5
&\bf 23.2 &\bf 7.9
&\bf 38.4 &\bf 12.0
&\bf 27.1 &\bf 10.0
\\

\bottomrule
\end{tabular}
\caption{
Ablation study on MSCOCO$_{36}$ and XM3600. By default, we dynamically substitute the model's vocab when new languages arrive, whereas setting (7) requires the accessibility of corpora of all languages to construct a task-shared vocab.
}
\label{tab:ablation_TEIR}
\end{table*}
%%%%%%%%%%%%%%

\subsection{Main Results}
Table~\ref{tab:main} provides retrieval results of different models. 
Specifically, joint-learning models CLL-CLIP and M-CLIP~\cite{carlsson2022MCLIP} respectively achieve the highest AR scores on MSCOCO$_{36}$ and XM3600. As the joint-learning setting covers all languages at the (pre-)training stage, its results can be regarded as the \emph{upper bound} of CL models. 
When learning different languages incrementally, all CL models experience different levels of forgetting. 
Notably, our TEIR can consistently boost all CL models across all metrics and datasets, e.g., with absolute improvements ranging from 3.7\% to 10.2\% in text-to-image AR on XM3600. 
The improved performance demonstrates the generality of TEIR across various CL and PEFT methods, proves the validity of our approach to maintaining acquired language skills, and highlights the importance of proper token embedding initialization and regularization.

%%%%%%%%%%%%%%
\begin{figure}[t]
\centering
\includegraphics[width=1\linewidth]{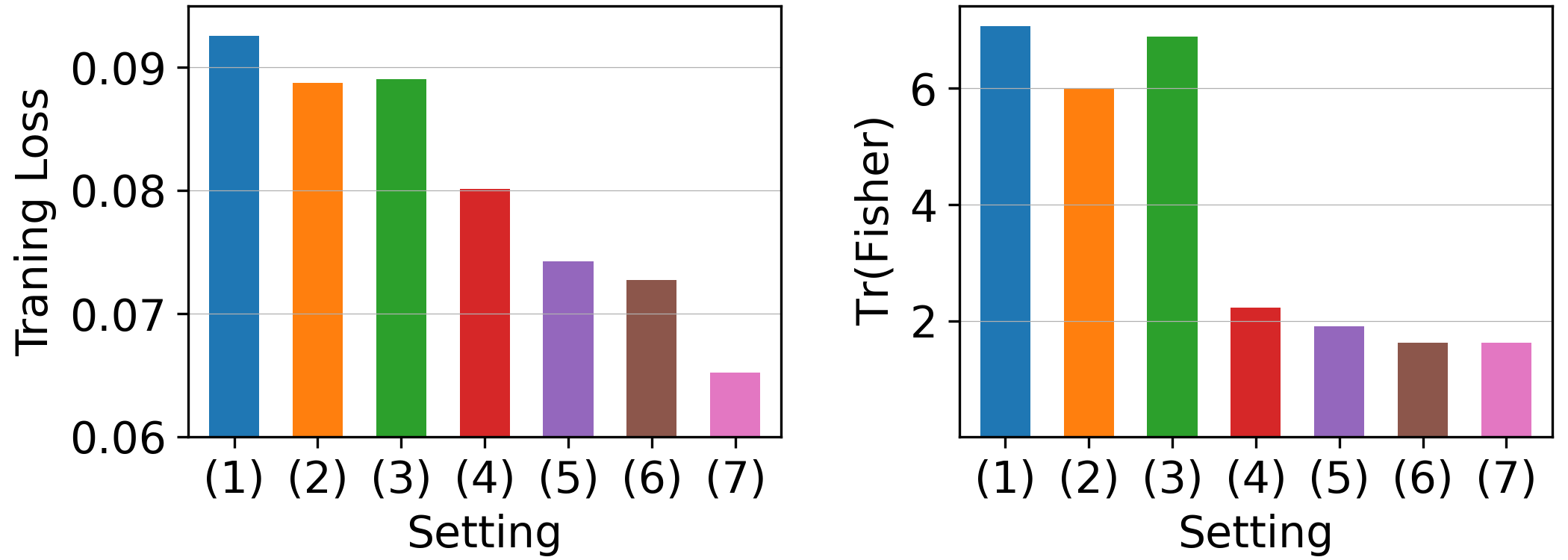}
\caption{Convergence analysis for different settings in Table~\ref{tab:ablation_TEIR}, focusing (left) the training loss and (right) the Fisher eigenvalues. Lower values respectively indicate closer to global minima and the convergence to flatter minima.
}
\label{fig:ablation_perturbation_fish}
\end{figure}
%%%%%%%%%%%%%%

\subsection{Ablations and Additional Analyses}
In the following, we delve deeper into our proposals via ablation studies and additional analyses, with ``CLL-CLIP with TEIR'' as the default model unless otherwise specified.

\subsubsection{Effect of Initialization} Table~\ref{tab:ablation_TEIR}(1,2) shows that ensuring identical distribution of new and prior token embeddings during initialization improves AR and F metrics by large margins. 
Compared with setting (5), setting (6) can still improve the model's learning capacity without sacrificing memory stability. 
These results suggest the importance of addressing the covariate shift problem in CLL. 

\subsubsection{Effect of Regularization} Table~\ref{tab:ablation_TEIR}(3,4) shows that imposing constraints on gradients or L2 weight decay when updating token embeddings can effectively mitigate the catastrophic forgetting problem of CLL-CLIP. So, it is crucial to penalize the embedding learning of lexically overlapping tokens and keep unrelated token embeddings intact. Moreover, the superiority of setting (5) against (3,4) indicates the complementary nature of these two strategies. Since our regularization method solely relies on the lexical statistics of the data, it incurs negligible additional costs, e.g., the training time of settings (1,6) is 11.1 and 11.3 hours, respectively.

\subsubsection{Effect of Vocab Substitution Strategy} We stick to the principle of continual learning and thus dynamically substitute the model's vocab when new languages arrive. In contrast, if we are allowed to access corpora of all languages, we can build an \emph{oracle} vocab and only need to substitute the model's vocab at the beginning. As shown in Table~\ref{tab:ablation_TEIR}(6,7), using the oracle vocab contributes to a boost in performance. Since we employ BPE to construct vocab in this work, the improvements confirm BPE's capacity to learn more accurate merging operations of sub-word units from extensive text statistics. Therefore, the exploration of refined vocab substitution strategies is a valuable avenue in future studies.

%%%%%%%%%%%%%%
\begin{figure}[t]
\centering
\includegraphics[width=1\linewidth]{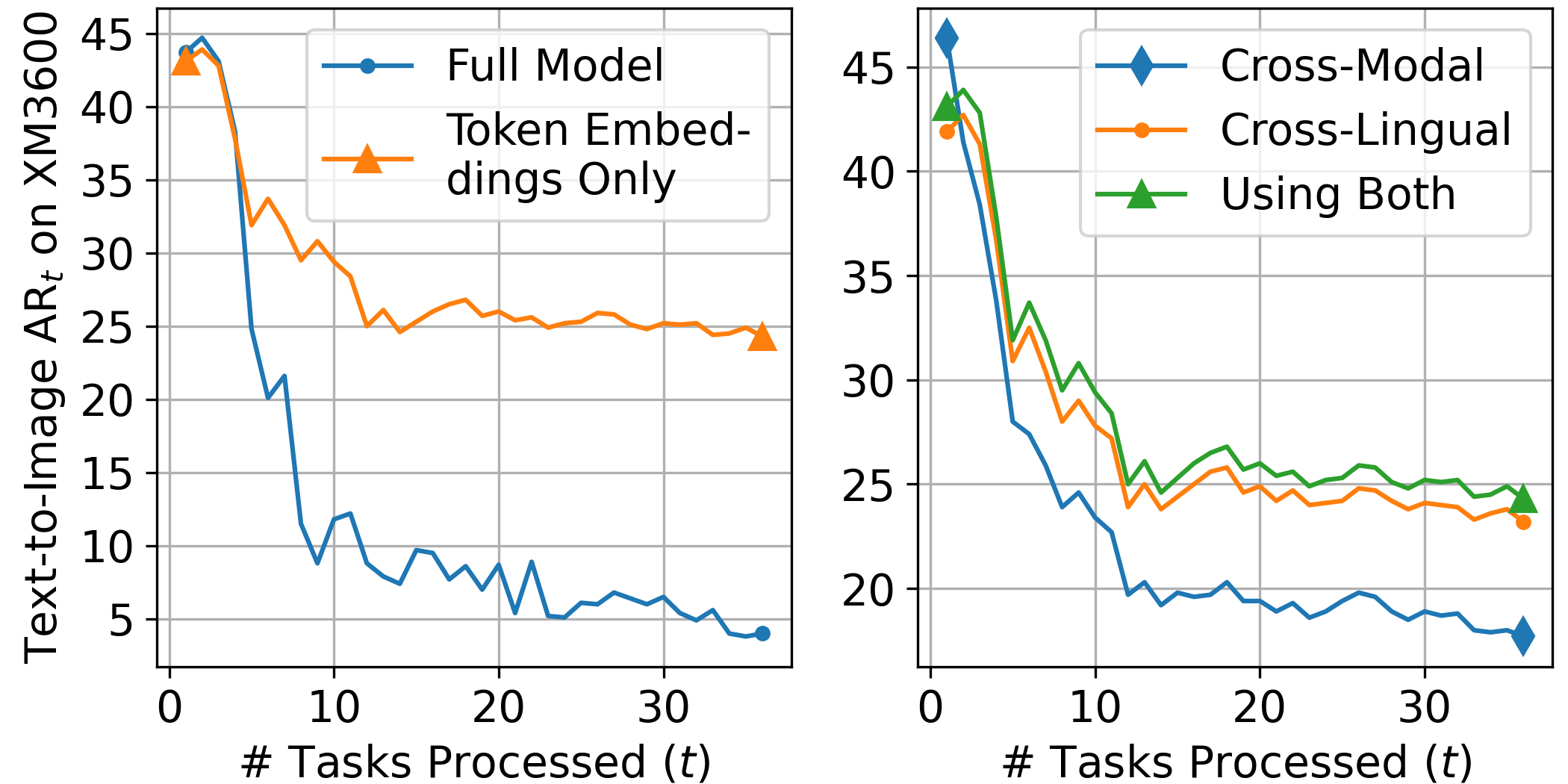}
\caption{Analysis of CLL-CLIP's core designs: (left) trainable components and (right) training objectives.
}
\label{fig:CLL_CLIP_analysis}
\end{figure}
%%%%%%%%%%%%%%

\subsubsection{Effect of TEIR on Model Convergence} 
We consider the model at the end of training and measure the property of the training minima of settings (1-7) in Table~\ref{tab:ablation_TEIR}. Firstly, we calculate the average loss across all training samples of MSCOCO$_{36}$. 
As depicted in Figure~\ref{fig:ablation_perturbation_fish}(left), the training loss of settings (2-7) is lower than that of (1), illustrating that TEIR facilitates the convergence of CLL-CLIP towards a \emph{global} minimum. Furthermore, we compute the trace of the empirical Fisher information matrix w.r.t. all training samples of MSCOCO$_{36}$ and treat it as a proxy for Hessian eigenvalues following \cite{chaudhari2017Entropy-SGD,kirkpatrick2017EWC,buzzega2020DER}. As depicted in Figure~\ref{fig:ablation_perturbation_fish}(right), settings (2-7) produce lower eigenvalues than (1), revealing that TEIR helps CLL-CLIP converge to a \emph{flatter} minimum.

\subsubsection{Effectiveness of CLL-CLIP} We here ablate the core designs of CLL-CLIP, including the trainable components and training objectives. As shown in Figure~\ref{fig:CLL_CLIP_analysis}(left), training the full model obtains dramatically degrades as more tasks are processed. Instead, our proposal of solely training token embeddings can preserve knowledge effectively. In Figure~\ref{fig:CLL_CLIP_analysis}(right), we can find that the cross-lingual objective is more efficient than the cross-modal objective to align images with multi-lingual texts, and leveraging both of them can achieve better results. This observation indicates the potential of utilizing text-only pairs for CLL.

%%%%%%%%%%%%%%
\begin{figure}[t]
\centering
\includegraphics[width=1\linewidth]{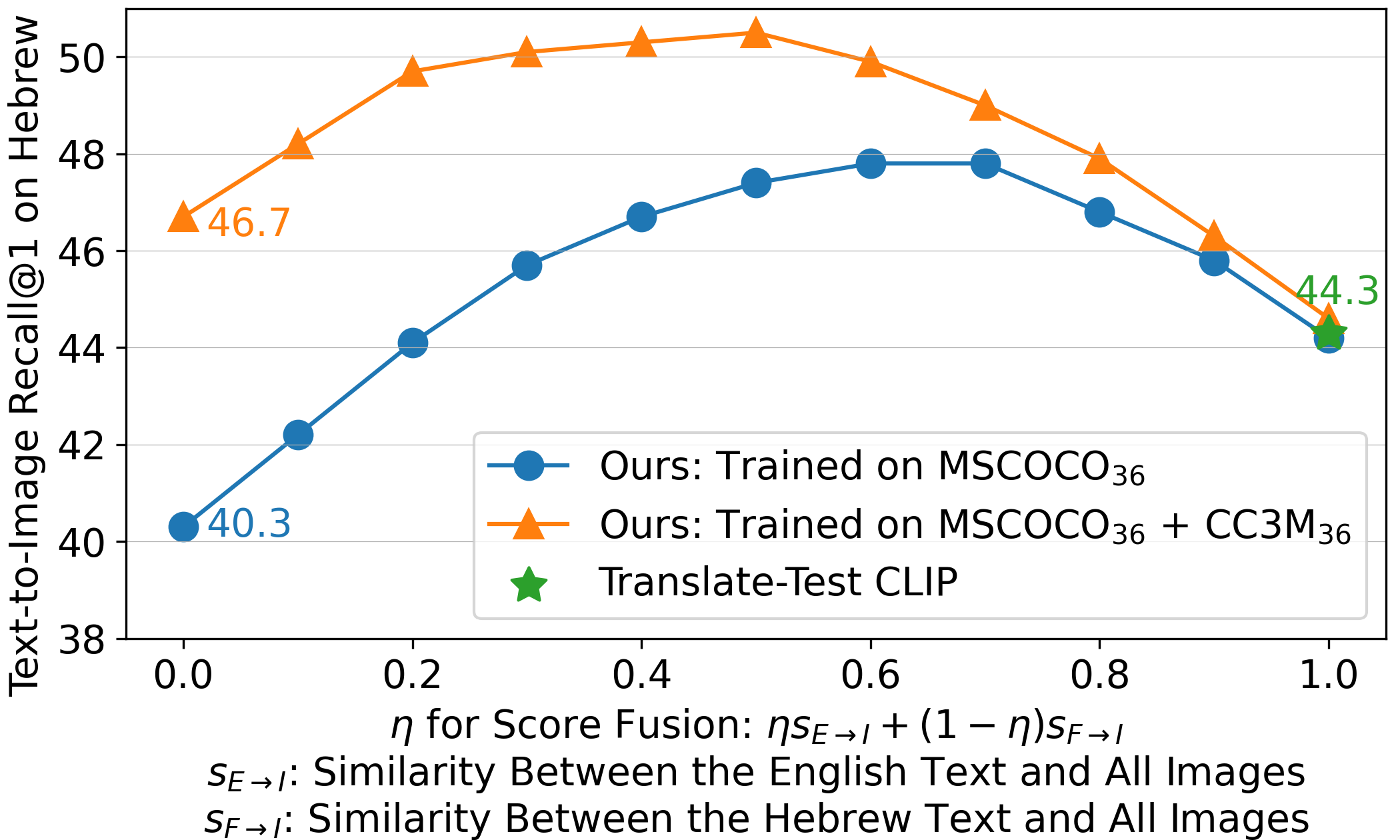}
\caption{Translate-test performance on Hebrew data in XM3600. Although translate-test CLIP is a strong pipeline system, our model can process foreign texts directly ($\eta = 0$) or achieve better retrieval performance via score fusion when translations are available ($\eta > 0$).
}
\label{fig:ablation_translate_test}
\end{figure}
%%%%%%%%%%%%%%

\subsubsection{Comparison with Translate-Test CLIP} To enable the original CLIP to understand multilingual texts, one intuitive approach is translating foreign texts into English, which is known as \emph{Translate Test} in the literature~\cite{bugliarello2022IGLUE,li2023WS-mVLP}. In Figure~\ref{fig:ablation_translate_test}, we compare our model with the translate-test CLIP under different $\eta$ for score fusion. Specifically, we can see that translate-test CLIP is a strong pipeline system that can achieve 44.3 text-to-image Recall@1 on Hebrew. In contrast, our CLL model can process Hebrew texts directly ($\eta = 0$). Encouragingly, our single model can even surpass the translate-test CLIP (46.7 vs. 44.3) when including an additional augmented CC3M dataset~\cite{sharma2018CC3M} for training. Therefore, continual language learning presents a viable avenue to evade the computation costs of translation and the error accumulation problem of a translation-based pipeline system. Moreover, when translations are available ($\eta > 0$), our model can simultaneously measure image-English text and image-Hebrew text similarities to achieve score fusion, leading to better retrieval performance compared with the case $\eta = 0$.

\subsubsection{Comparisons with Multilingual VL-PTMs} In Table~\ref{tab:multi30k}, we compare CLL models with multilingual VL-PTMs on Multi30K~\cite{elliott2016Multi30K}. As we can see, although the joint-learning MLA model performs generally the best among the four languages, it obtains inferior performance in Czech compared with ``MLA with TEIR'' which has learned 36 languages in a continual learning manner. Given the gap between joint learning and continual learning, there is much room for improving CLL models.

%%%%%%%%%%%%%%%%%%%%%%
\begin{table}[tbp]
\centering
% \tabfootnotesize

\setlength{\tabcolsep}{2.5pt}
% \fontsize{8}{9}\selectfont
\begin{tabular}{@{}llccccc@{}}
\toprule 

Setting &Model &en &de &fr &cs & Avg.
\\
\midrule

\small{Learn in English} &CLIP {\small \citeyearpar{radford2021CLIP}} &86.3 &38.4 &48.9 &8.1 &45.4\\
\midrule

\multirow{3}{*}{\begin{tabular}[c]{@{}l@{}}Joint Learning\\ \small($<$10 languages)\end{tabular}}

&M$^3$P {\small \citeyearpar{ni2021M3P}} &57.9 &36.8 &27.1 &20.4 &35.6\\
&UC$^2$ {\small \citeyearpar{zhou2021UC2}} &66.6 &62.5 &60.4 &55.1 &61.2\\
&MLA {\small \citeyearpar{zhang2022MLA}} &\bf 86.4 &\bf 80.8 &\bf 80.9 &72.9 &\bf 80.3\\
\cmidrule{2-7}

% \rowcolor{gray!10} 
\small($>$60 languages) &M-CLIP {\small \citeyearpar{carlsson2022MCLIP}} &84.1 &79.1 &77.5 &\bf 76.3 &79.3\\

\midrule

\multirow{4}{*}{\begin{tabular}[c]{@{}l@{}}Continual\\Learning\\ \small(36 languages)\end{tabular}}

&CLL-CLIP &75.1 &36.2 &46.5 &57.6 &53.8\\
&\quad\bf with TEIR &\bf 82.5 &\bf 48.6 &\bf 60.4 &\bf 66.5 &\bf 64.5\\

\cmidrule{2-7}

&MLA {\small \citeyearpar{zhang2022MLA}} &73.8 &42.7 &52.7 &64.6 &58.4\\
&\quad\bf with TEIR &\bf 82.4 &\bf 58.1 &\bf 68.0 &\bf 74.1 &\bf 70.7\\

\bottomrule
\end{tabular}

\caption{Zero-shot image-text retrieval results (averaged over recall@\{1,5,10\} on two directions) on Multi30K under English (en), German (de), French (fr), and Czech (cs).
}
\label{tab:multi30k}
\end{table}
%%%%%%%%%%%%%%%%%%%%%%

\section{Conclusion}
In this paper, we present to our best knowledge the first systematical study on extending the language capacities of dual-stream vision-language pre-trained models (VL-PTMs) under the practical continual language learning setting. We introduce a CLL-CLIP model and a TEIR approach to learn the alignment between images and multilingual texts while mitigating catastrophic forgetting raised by the covariate shift and lexical overlap problems. To comprehensively validate our proposals, we construct a benchmark spanning 36 languages and conduct evaluations on multilingual image-text retrieval. Through a series of experiments and analyses, we verify the effectiveness of CLL-CLIP and TEIR and gain insights into their inner workings. We hope our research can serve as a basis to enhance the accessibility of VL-PTMs across different linguistic communities.

\paragraph{Limitations} This paper focuses exclusively on the continual language learning of CLIP-like VL-PTMs, emphasizing evaluations for image-text retrieval. Nonetheless, we posit that our ideas hold the potential to be adaptable to encoder-decoder-based VL-PTMs and generation tasks like visual captioning~\cite{yang2023CARE}. We leave it to our future study. Moreover, TEIR requires current-task text statistics to compute Equation~(\ref{eq:coeff}), making it difficult to handle the challenges posed by, e.g., \emph{boundary-free continual learning}~\cite{aljundi2019taskfree}.

\section*{Acknowledgements} {This paper was partially supported by NSFC (No. 6217600\\8), the project of Pengcheng Laboratory (PCL2023A08), and Shenzhen Science and Technology Research Program (No. GXWD20201231165807007-20200814115301001)}.

\bibliography{aaai24}

%%%%%%%%%%%%%%%%%%%%%%
\begin{table}[htbp]
\rowcolors{2}{gray!15}{white}
\setlength\tabcolsep{2pt}
\centering
\fontsize{8}{9}\selectfont
\begin{tabular}{ccccc}
\toprule
\begin{tabular}[c]{@{}c@{}}ISO\\Code\end{tabular} &Language &Script &Family &Branch\\
\midrule
ar	&Arabic	&Arabic	&Afro-Asiatic	&\\
bn	&Bengali	&Bengali	&Indo-European	&Indo-Iranian\\
cs	&Czech	&Latin	&Indo-European	&Balto-Slavic\\
da	&Danish	&Latin	&Indo-European	&North Germanic\\
de	&German	&Latin	&Indo-European	&West Germanic\\
el	&Greek	&Latin	&Indo-European	&Hellenic\\
en	&English	&Latin	&Indo-European	&West Germanic\\
es	&Spanish	&Latin	&Indo-European	&Italic\\
fa	&Persian	&Arabic	&Indo-European	&Indo-Iranian\\
fi	&Finnish	&Latin	&Uralic	&Finnic\\
fil	&Filipino	&Latin	&Austronesian	&Malayo-Polynesian\\
fr	&French	&Latin	&Indo-European	&Italic\\
he	&Hebrew	&Hebrew	&Afro-Asiatic	&Semitic\\
hi	&Hindi	&Devanagari	&Indo-European	&Indo-Iranian\\
hr	&Croatian	&Latin	&Indo-European	&Balto-Slavic\\
hu	&Hungarian	&Latin	&Uralic	&\\
id	&Indonesian	&Latin	&Austronesian	&Malayo-Polynesian\\
it	&Italian	&Latin	&Indo-European	&Italic\\
ja	&Japanese	&Kanji	&Japonic	&\\
ko	&Korean	&Hangul	&Koreanic	&\\
mi	&Māori	&Latin	&Austronesian	&Malayo-Polynesian\\
nl	&Dutch	&Latin	&Indo-European	&West Germanic\\
no	&Norwegian	&Latin	&Indo-European	&North Germanic\\
pl	&Polish	&Latin	&Indo-European	&Balto-Slavic\\
pt	&Portuguese	&Latin	&Indo-European	&Italic\\
quz	&Cuzco Quechua	&Latin	&Quechuan	&\\
ro	&Romanian	&Latin	&Indo-European	&Italic\\
ru	&Russian	&Cyrillic	&Indo-European	&Balto-Slavic\\
sv	&Swedish	&Latin	&Indo-European	&North Germanic\\
sw	&Swahili	&Latin	&Niger–Congo	&\\
te	&Telugu	&Telugu	&Dravidian	&South-Central\\
th	&Thai	&Thai	&Kra-Dai	&Tai\\
tr	&Turkish	&Latin	&Turkic	&\\
uk	&Ukrainian	&Cyrillic	&Indo-European	&Balto-Slavic\\
vi	&Vietnamese	&Latin	&Austroasiatic	&\\
zh	&Chinese	&\begin{tabular}[c]{@{}c@{}}Chinese\\Characters\end{tabular} &Sino-Tibetan	&Sinitic\\
\bottomrule
\end{tabular}
\caption{Information of the 36 languages considered for continual language learning.}
\label{tab:lang_info}
\end{table}
%%%%%%%%%%%%%%%%%%%%%%

\section*{Appendix}

\subsection*{Language Details and Per-Language Statistics}
As shown in Table~\ref{tab:lang_info}, the 36 languages considered for CLL have different scripts, families, or branches, leading to great challenges in our problem setup. Next, we give per-language statistics of MSCOCO$_{36}$ and XM3600 datasets in Table~\ref{tab:language_statistics}. As we can see, the number of words and characters per caption varies significantly in different languages. These results indicate high linguistic differences, which pose great challenges to building a single vision-language model that can align visual information with texts in the 36 languages.

%%%%%%%%%%%%%%%%%%%%%%
\begin{table*}[htbp]
\rowcolors{2}{gray!15}{white}
\centering
\begin{tabular}{c ccc ccc}
\toprule 
&\multicolumn{3}{c}{MSCOCO$_{36}$}
&\multicolumn{3}{c}{XM3600}\\
\cmidrule(lr){2-4}
\cmidrule(lr){5-7}
\multirow{-2}{*}{ISO Code}
&\# Captions 
&\begin{tabular}[c]{@{}c@{}}\# Words\\per Caption\end{tabular} 
&\begin{tabular}[c]{@{}c@{}}\# Chars\\per Caption\end{tabular} 
&\# Captions 
&\begin{tabular}[c]{@{}c@{}}\# Words\\per Caption\end{tabular} 
&\begin{tabular}[c]{@{}c@{}}\# Chars\\per Caption\end{tabular} 
\\
\midrule
ar	&616,767	&7.6	&41.2	&7,367	&7.7	&42.2\\
bn	&616,767	&8.6	&50.9	&3,600	&11.3	&62.1\\
cs	&616,767	&7.6	&45.9	&7,207	&6.5	&39.1\\
da	&616,767	&9.6	&51.6	&7,264	&8.7	&48.3\\
de	&616,767	&9.5	&60.3	&8,643	&11.2	&76.5\\
el	&616,767	&10.2	&61.9	&7,204	&7.7	&51.4\\
en	&616,767	&10.5	&52.4	&7,200	&9.4	&49.5\\
es	&616,767	&11.1	&59.9	&8,614	&9.8	&56.3\\
fa	&616,767	&10.9	&48.9	&7,245	&12.7	&59.4\\
fi	&616,767	&6.1	&50.7	&7,127	&7.5	&65.2\\
fil	&616,767	&11.7	&66.6	&7,109	&12.2	&67.6\\
fr	&616,767	&10.9	&60.7	&8,562	&12.3	&69.6\\
he	&616,767	&7.0	&36.6	&7,200	&11.9	&63.6\\
hi	&616,767	&10.9	&50.4	&8,503	&13.4	&59.9\\
hr	&616,767	&7.8	&47.4	&7,280	&9.0	&57.8\\
hu	&616,767	&7.7	&49.8	&7,216	&8.5	&60.5\\
id	&616,767	&8.7	&57.0	&7,126	&14.3	&93.5\\
it	&616,767	&10.7	&60.0	&8,471	&12.1	&71.8\\
ja	&616,767	&1.3	&22.2	&7,185	&1.0	&26.0\\
ko	&616,767	&6.9	&24.9	&7,650	&7.0	&24.7\\
mi	&616,767	&13.7	&61.6	&4,732	&11.7	&55.5\\
nl	&616,767	&9.8	&56.0	&8,059	&8.0	&45.9\\
no	&616,767	&9.6	&51.4	&7,213	&9.6	&54.3\\
pl	&616,767	&7.6	&50.7	&7,141	&8.3	&57.6\\
pt	&616,767	&10.6	&57.4	&7,243	&10.8	&61.7\\
quz	&616,767	&7.0	&55.6	&7,200	&5.0	&38.6\\
ro	&616,767	&10.4	&55.4	&7,123	&15.6	&88.4\\
ru	&616,767	&7.5	&49.5	&7,200	&9.9	&66.3\\
sv	&616,767	&9.4	&51.6	&7,273	&8.1	&46.7\\
sw	&616,767	&9.0	&56.4	&7,046	&10.7	&63.0\\
te	&616,767	&7.3	&53.2	&7,200	&7.1	&47.4\\
th	&616,767	&1.2	&38.7	&7,200	&1.2	&47.9\\
tr	&616,767	&8.0	&52.2	&7,233	&9.4	&63.4\\
uk	&616,767	&7.5	&49.4	&7,215	&10.0	&65.7\\
vi	&616,767	&12.7	&56.1	&7,350	&18.0	&79.3\\
zh	&616,767	&1.0	&15.6	&7,174	&1.0	&23.0\\
\bottomrule
\end{tabular}
\caption{Per-language statistics. Besides the total number of captions, we also report the number of words (where applicable) and the number of characters per sentence. Note that we split the caption based on white space to calculate the number of words, which does not apply to languages without boundaries, e.g., Japanese (ja), Korean (ko), Thai (the), and Chinese (zh).}
\label{tab:language_statistics}
\end{table*}
%%%%%%%%%%%%%%%%%%%%%%

%%%%%%%%%%%%%%%%%%%%%%
\begin{figure}[htbp]
\centering
\includegraphics[width=0.88\linewidth]{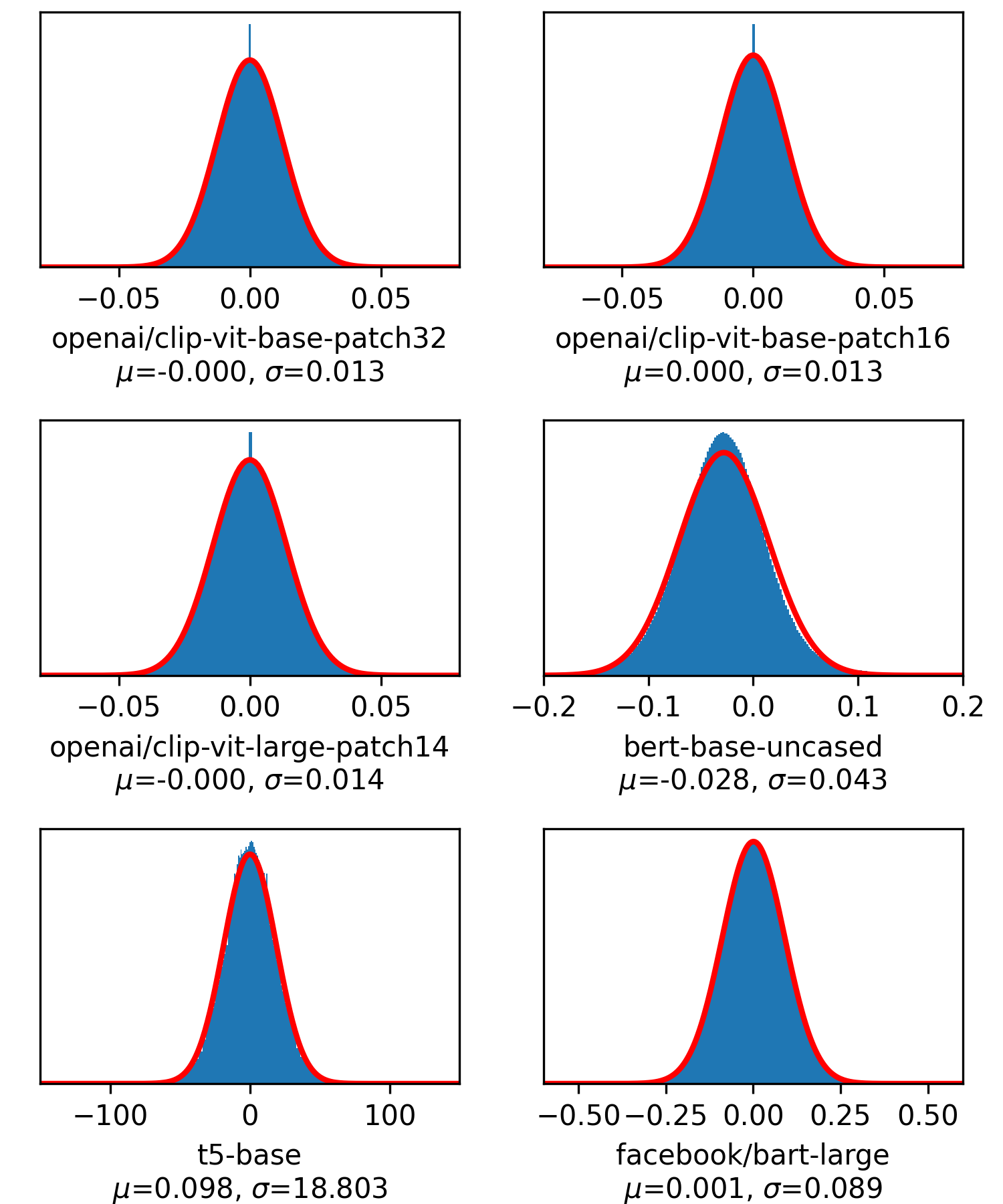}
\caption{Token embedding distribution of different pre-trained models (PTMs). Below each subfigure, we give the PTM name in Hugging Face and the mean ($\mu$) and standard deviation ($\sigma$) of its token embeddings. Gaussian distribution of the same $\mu$ and $\sigma$ is illustrated in red lines.}
\label{fig:ted_pretrained}
\end{figure}
%%%%%%%%%%%%%%%%%%%%%%

%%%%%%%%%%%%%%%%%%%%
\begin{figure*}[tbp]
\centering
\includegraphics[width=0.88\linewidth]{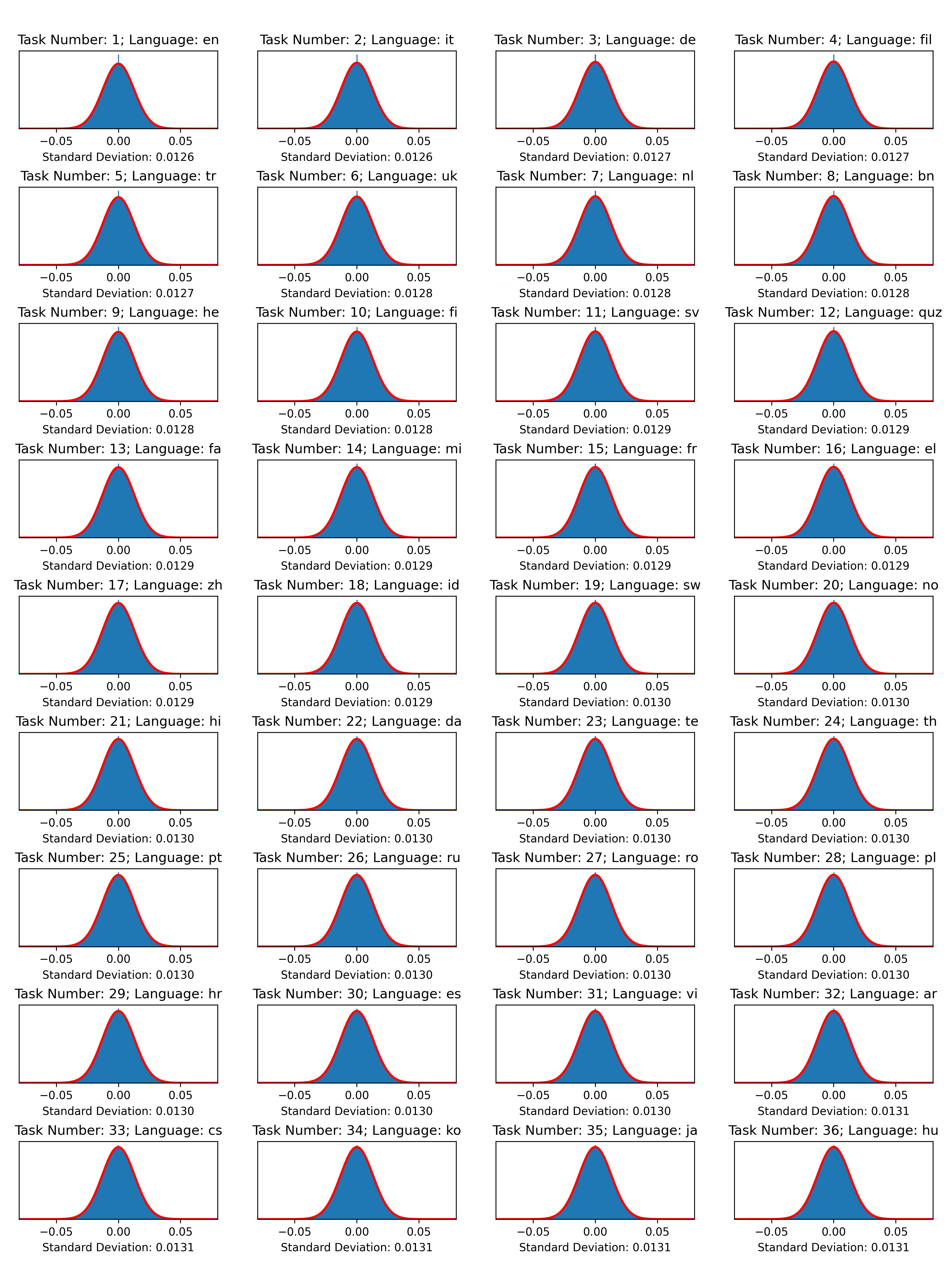}
\caption{Token embedding distribution (TED) of CLL-CLIP with TEIR after learning every task. New token embeddings are initialized with the identical distribution of previously learned token embeddings. As we can observe, the TED of this model can fit the Gaussian distribution of the same standard deviation (red lines) well, no matter how many tasks have been processed. Therefore, our assumption $\bm{\theta}^*_{t-1} \sim \mathcal{N}(\mu_{t-1}, \sigma_{t-1}^2)$ generally holds if the initialization of $\bm{\theta}_{t-1}$ follows a Gaussian distribution.
}
\label{fig:ted_cll_clip_with_teir}
\end{figure*}
%%%%%%%%%%%%%%%%%%%%

\subsection*{Token Embedding Distribution of PTMs}
As stated in the paper, language models building on Transformer \cite{vaswani2017Transformer} typically initialize token embeddings with a Gaussian distribution $\mathcal{N}(\mu, \sigma^2)$ with zero mean ($\mu=0$) and a pre-defined variance $\sigma^2$. Here, we first examine the token embedding distribution (TED) of representative PTMs. We consider text encoders of different variants of CLIP~\cite{radford2021CLIP}, BERT~\cite{devlin2019BERT}, T5~\cite{raffel2020T5}, and BART~\cite{lewis2020BART}. As shown in Figure~\ref{fig:ted_pretrained}, TED of different PTMs fit Gaussian distributions well, indicating token embeddings initialized with $\mathcal{N}(\mu, \sigma^2)$ generally follow a Gaussian-like distribution after training. Next, we examine how TED of our ``CLL-CLIP with TEIR'' changes during continual language learning. We can observe in Figure~\ref{fig:ted_cll_clip_with_teir} that no matter how many tasks have been processed, the TED of our model fits Gaussian distributions well. 

Based on the above results, we empirically verify that our assumption, i.e., $\bm{\theta}^*_{t-1} \sim \mathcal{N}(\mu_{t-1}, \sigma_{t-1}^2)$, generally holds if the initialization of $\bm{\theta}_{t-1}$ follows a Gaussian distribution.

%%%%%%%%%%%%%%%%%%%%
\begin{table*}[htbp]
\rowcolors{2}{gray!15}{white}
\centering
\begin{tabular}{lcc}
\toprule
&Order &Basis\\
\midrule
(1)
&\begin{tabular}[c]{@{}c@{}}
ar, bn, cs, da, de, el, en, es, fa, fi, fil, fr,\\
he, hi, hr, hu, id, it, ja, ko, mi, nl, no, pl,\\
pt, quz, ro, ru, sv, sw, te, th, tr, uk, vi, zh
\end{tabular}
&alphabetical order of ISO codes\\

(2)
&\begin{tabular}[c]{@{}c@{}}
it, de, fil, tr, uk, nl, bn, he, fi, sv, quz, fa,\\
mi, fr, el, zh, id, sw, no, hi, da, te, th, pt,\\
ru, ro, pl, hr, es, vi, ar, cs, ko, ja, hu, en
\end{tabular}
&shuffle (1) by the random seed 222\\

(3): default
&\begin{tabular}[c]{@{}c@{}}
en, it, de, fil, tr, uk, nl, bn, he, fi, sv, quz,\\
fa, mi, fr, el, zh, id, sw, no, hi, da, te, th\\
pt, ru, ro, pl, hr, es, vi, ar, cs, ko, ja, hu
\end{tabular}
&based on (2); English places the first\\

\bottomrule
\end{tabular}
\caption{Task order variants. The random seed 222 is randomly selected. Order (3) is our default choice in the paper. We train VL-PTMs on the English task first to obtain specialist models and then extend them to other 35 languages.}
\label{tab:order}
\end{table*}

\begin{table*}[htbp]
\centering
% \tabfootnotesize
\setlength{\tabcolsep}{2pt}
\small
% \fontsize{8}{9}\selectfont
\begin{tabular}{@{}l llll llll llll@{}}
\toprule 
\multirow{3}{*}[-5pt]{Model} 
&\multicolumn{4}{c}{Task Order (1)}
&\multicolumn{4}{c}{Task Order (2)}
&\multicolumn{4}{c}{Task Order (3)}
\\
\cmidrule(lr){2-5} 
\cmidrule(lr){6-9}
\cmidrule(lr){10-13}
&\multicolumn{2}{c}{Image-to-Text}
&\multicolumn{2}{c}{Text-to-Image}
&\multicolumn{2}{c}{Image-to-Text}
&\multicolumn{2}{c}{Text-to-Image}
&\multicolumn{2}{c}{Image-to-Text}
&\multicolumn{2}{c}{Text-to-Image}
\\ 
\cmidrule(lr){2-3} 
\cmidrule(lr){4-5}
\cmidrule(lr){6-7} 
\cmidrule(lr){8-9}
\cmidrule(lr){10-11} 
\cmidrule(lr){12-13}
&AR ($\uparrow$) &F ($\downarrow$) 
&AR ($\uparrow$) &F ($\downarrow$)
&AR ($\uparrow$) &F ($\downarrow$) 
&AR ($\uparrow$) &F ($\downarrow$)
&AR ($\uparrow$) &F ($\downarrow$) 
&AR ($\uparrow$) &F ($\downarrow$)
\\
\midrule% [\heavyrulewidth]  

ER \citeyearpar{chaudhry2019ER} 
&28.3 &20.5
&19.3 &16.2
&28.0 &20.9
&19.0 &16.2
&29.0 &20.0
&19.4 &16.0
\\
\quad \bf with TEIR
&\bf 35.6 \tiny (+7.3)
&\bf 13.8 \tiny (+6.7)
&\bf 25.0 \tiny (+5.7)
&\bf 11.1 \tiny (+5.1)
&\bf 35.0 \tiny (+7.0) 
&\bf 14.2 \tiny (+6.7)
&\bf 24.5 \tiny (+5.5)
&\bf 11.4 \tiny (+4.8)
&\bf 35.4 \tiny (+6.4) 
&\bf 13.9 \tiny (+6.1)
&\bf 24.7 \tiny (+5.3)
&\bf 11.2 \tiny (+4.8)
\\
\midrule

DER \citeyearpar{buzzega2020DER}
&31.4 &17.6
&20.9 &14.6
&30.8 &18.3
&20.7 &14.6
&31.6 &17.4
&21.0 &14.4
\\
\quad \bf with TEIR
&\bf 37.5 \tiny (+6.1) 
&\bf 11.8 \tiny (+5.8) 
&\bf 26.3 \tiny (+5.4) 
&\bf 9.6 \tiny (+5.0) 
&\bf 38.1 \tiny (+7.3) 
&\bf 11.1 \tiny (+7.2) 
&\bf 26.7 \tiny (+6.0) 
&\bf 9.2 \tiny (+5.4) 
&\bf 38.3 \tiny (+6.7) 
&\bf 10.9 \tiny (+6.5)
&\bf 26.7 \tiny (+5.7) 
&\bf 9.3 \tiny (+5.1)
\\
\midrule

MLA \citeyearpar{zhang2022MLA} 
&31.4 &21.1
&20.9 &17.8
&30.7 &21.8
&20.4 &18.4
&30.7 &21.8
&20.6 &18.1
\\

\quad \bf with TEIR
&\bf 41.1 \tiny (+9.7)
&\bf 12.4 \tiny (+8.7)
&\bf 29.2 \tiny (+8.3)
&\bf 10.4 \tiny (+7.4)
&\bf 41.3 \tiny (+10.6)
&\bf 12.2 \tiny (+9.6)
&\bf 29.2 \tiny (+8.8)
&\bf 10.5 \tiny (+7.9)
&\bf 41.1 \tiny (+10.4)
&\bf 12.3 \tiny (+9.5)
&\bf 29.0 \tiny (+8.4)
&\bf 10.7 \tiny (+7.4)
\\

\bottomrule
\end{tabular}
\caption{Retrieval performance on XM3600 under different task orders listed in Table~\ref{tab:order}. The numbers in brackets indicate the absolute improvements brought by our approach. We can see that models trained in different task orders obtain similar results in the end, and our proposed TEIR can consistently improve various models. Full results of the task order (2) are given in Table~\ref{tab:order2_full}.}
\label{tab:order_results}
\end{table*}
%%%%%%%%%%%%%%%%%%%%

\subsection*{Task Order}
In this work, we decide the task order by a random seed 222 and make the English language place the first, as shown in Table~\ref{tab:order}. To evaluate the impact of different task orders, we consider three order variants listed in Table~\ref{tab:order} and compare six models, i.e., ER~\cite{chaudhry2019ER}, DER~\cite{buzzega2020DER}, and MLA~\cite{zhang2022MLA} with or without our proposed TEIR. From Table~\ref{tab:order_results}, we can see that (1) models trained in different task orders obtain similar results in the end; (2) our proposed TEIR can consistently improve various models under different task orders.

\subsection*{Reproducibility}
Our code and data are available at \url{https://github.com/yangbang18/CLFM}. Our code includes the re-production of all SOTA methods, with details described below.
\begin{itemize}
    \item {\bf M-CLIP}~\cite{carlsson2022MCLIP} is a PTM trained in 69 languages via a joint-learning setup. We directly test it on MSCOCO$_{36}$ and XM3600 without re-training.
    
    \item {\bf oEWC}~\cite{schwarz2018oEWC} penalizes the changes in the trainable token embeddings of CLL-CLIP by adding a regularization item based on the (diagonal) Fisher information matrix to the loss function. So the overall loss now becomes $\mathcal{L} + \frac{\lambda}{2}||\theta_t-\theta^*_{t-1}||^2_{\gamma F^*_{t-1}}$ (please see its paper for explanations). Here are two hyper-parameters: the strength factor $\lambda$ and the accumulative factor $\gamma$. We empirically set $\gamma=1.0$ and search $\lambda$ from $\{1, 10, 100, 1000\}$ and finally set $\lambda=1000$ based on the validation performance.
    
    \item {\bf ER}~\cite{chaudhry2019ER} stores historical training samples for current-task learning of CLL-CLIP. It has a buffer size hyper-parameter and we empirically set it to 8,000. Following its paper, we use the reservoir sampling strategy to store triplet samples. With a memory dataset $M$ and a training dataset $D_t$, we separately sample a batch of data from $M$ and $D_t$ to calculate loss $\mathcal{L}$ at each training iteration. 
    
    \item {\bf DER}~\cite{buzzega2020DER} stores features of previously learned samples for distilling knowledge to CLL-CLIP. Thus, the overall loss function becomes $\mathcal{L} + \alpha \mathcal{L}_{\text{KD}}$. We set $\alpha = 1$ empirically and set the buffer size to 8,000 (identical to ER). Moreover, we store features of historical foreign texts into a memory $M$ and let $\mathcal{L}_{\text{KD}} = \frac{1}{2K} \sum_{k=1}^K ||\bm{r}^M_k - \bm{r}^F_k||_2^2$, where $\bm{r}^M$ is sampled from $M$.
    
    \item {\bf MLA}~\cite{zhang2022MLA} inserts task-specific adapters~\cite{houlsby2019Adapter} into the frozen text encoder of CLL-CLIP. Following its paper, adapters are inserted after the MLP layers of Transformer blocks. Each adapter is implemented as a bottleneck MLP (one hidden layer) with a reduction factor of 2 and the ReLU non-linearity function.
    
    \item {\bf P-Tuning}~\cite{liu2022P-Tuning} inserts task-specific layer-wise learnable prompt tokens into the frozen text encoder of CLL-CLIP. We use the ``prefix\_tuning'' config in AdapterHub~\cite{pfeiffer2020AdapterHub}. Specifically, the number of prompt tokens per layer is 30, and a MLA (one hidden layer, hidden size 512) with the Tanh non-linearity function is used to learn these prompt tokens.
    
    \item {\bf LoRA}~\cite{hu2022LoRA} inserts low-rank matrices into the frozen text encoder of CLL-CLIP to calibrate attention. We use the ``lora'' config in AdapterHub~\cite{pfeiffer2020AdapterHub}. Specifically, LoRA layers are added to the key and value self-attention matrices, and the rank and the scaling factor of LoRA layers are 8.
    
    \item {\bf DualPrompt}~\cite{wang2022DualPrompt} uses a key-query mechanism to generate proper prompts for the frozen text encoder of CLL-CLIP. Following its paper, we insert G(eneral)-Prompts into the first two layers and insert E(xpert)-Prompts into the next three layers. We set the pool size of E-Prompts to the number of tasks (i.e., 36). Moreover, the method has three more hyper-parameters, the number E-Prompts to be matched ($k$), the length of G-Prompts $l_g$ and the length of E-Prompts $l_e$. We search $k$ from $[1,5]$, $l_g$ and $l_e$ from $\{2, 4, 8, 16\}$. We set $k=1$, $l_g=2$, and $l_e=16$ based on validation performance.
    
    \item {\bf CodaPrompt}~\cite{smith2023CodaPrompt} shares the same spirit as DualPrompt, but the core difference is that it uses an attention-based end-to-end key-query scheme. Following its paper, we insert E(xpert)-Prompts into the first five layers of the frozen text encoder of CLL-CLIP. The method has two more hyper-parameters: the length of E-Prompts $l_e$ and the pool size of E-Prompts $p_e$. We search $l_e$ from $\{2, 4, 8, 16\}$ and $p_e$ from $\{36\times1, 36\times3, 36\times5, 36\times7\}$. Based on validation performance, we set $l_e=2$, $p_e=36\times5$.
\end{itemize}

%%%%%%%%%%%%%%%%
\begin{table*}[htbp]
\centering
\setlength{\tabcolsep}{3pt}
% \fontsize{8}{9}\selectfont
\begin{tabular}{@{}ll llll llll@{}}
\toprule 
\multirow{3}{*}[-5pt]{\begin{tabular}[c]{@{}c@{}}Setting\end{tabular}} 
&\multirow{3}{*}[-5pt]{Model} 
&\multicolumn{4}{c}{MSCOCO$_{36}$ (In-Domain)}
&\multicolumn{4}{c}{XM3600 (Out-of-Domain)}
\\
\cmidrule(lr){3-6} 
\cmidrule(lr){7-10}
&
&\multicolumn{2}{c}{Image-to-Text}
&\multicolumn{2}{c}{Text-to-Image}
&\multicolumn{2}{c}{Image-to-Text}
&\multicolumn{2}{c}{Text-to-Image}
\\ 
\cmidrule(lr){3-4} 
\cmidrule(lr){5-6}
\cmidrule(lr){7-8} 
\cmidrule(lr){9-10}
&
&AR ($\uparrow$) &F ($\downarrow$) 
&AR ($\uparrow$) &F ($\downarrow$)
&AR ($\uparrow$) &F ($\downarrow$) 
&AR ($\uparrow$) &F ($\downarrow$)
\\
\midrule% [\heavyrulewidth] 

\multirow{3}{*}{\begin{tabular}[c]{@{}c@{}}Joint\\Training\end{tabular}}

&CLL-CLIP
&\bf 53.4 &-
&\bf 31.4 &-
&50.6 &-
&37.1 &-
\\

&M-CLIP~\citeyearpar{carlsson2022MCLIP}
&42.7 &-
&25.9 &-
&\bf 53.6 &-
&\bf 41.1 &-
\\

&PaLI~\citeyearpar{chen2023PaLI}
&- &-
&- &-
&36.0 &-
&28.5 &-
\\
\midrule

\multirow{18}{*}{\begin{tabular}[c]{@{}c@{}}Continual\\Learning\end{tabular}}
&CLL-CLIP
&25.2 &27.9
&13.0 &17.8
&21.9 &27.7
&14.6 &21.5
\\

&\quad \bf with TEIR
&\bf 37.6 \scriptsize (+12.4) 
&\bf 15.3 \scriptsize (+12.6)
&\bf 20.3 \scriptsize (+7.3)
&\bf 10.7 \scriptsize (+7.1)
&\bf 34.9 \scriptsize (+13.0) 
&\bf 15.3 \scriptsize (+12.4)
&\bf 24.4 \scriptsize (+9.8)
&\bf 12.4 \scriptsize (+9.1)
\\
\cmidrule(lr){2-10}

&oEWC \citeyearpar{schwarz2018oEWC}
&36.8 &15.9 &19.6 &11.1
&32.2 &17.2 &22.3 &13.8
\\
&\quad \bf with TEIR
&\bf 39.8 \scriptsize (+3.0)
&\bf 13.1 \scriptsize (+2.8)
&\bf 21.5 \scriptsize (+1.9)
&\bf 9.4 \scriptsize (+1.7)
&\bf 36.7 \scriptsize (+4.5)
&\bf 13.3 \scriptsize (+3.9)
&\bf 25.5 \scriptsize (+3.2)
&\bf 11.3 \scriptsize (+2.5)
\\
\cmidrule(lr){2-10}

&ER \citeyearpar{chaudhry2019ER}
&33.5 &18.5
&17.8 &12.3
&28.0 &20.9
&19.0 &16.2
\\
&\quad \bf with TEIR
&\bf 38.6 \scriptsize (+5.1) 
&\bf 13.4 \scriptsize (+5.1)
&\bf 21.4 \scriptsize (+3.6)
&\bf 8.9 \scriptsize (+3.4)
&\bf 35.0 \scriptsize (+7.0) 
&\bf 14.2 \scriptsize (+6.7)
&\bf 24.5 \scriptsize (+5.5)
&\bf 11.4 \scriptsize (+4.8)
\\
\cmidrule(lr){2-10}

&DER \citeyearpar{buzzega2020DER}
&37.0 &15.1
&19.5 &10.7
&30.8 &18.3
&20.7 &14.6
\\
&\quad \bf with TEIR
&\bf 42.6 \scriptsize (+5.6) 
&\bf 9.6 \scriptsize (+5.5) 
&\bf 23.5 \scriptsize (+4.0) 
&\bf 6.9 \scriptsize (+3.8) 
&\bf 38.1 \scriptsize (+7.3) 
&\bf 11.1 \scriptsize (+7.2) 
&\bf 26.7 \scriptsize (+6.0) 
&\bf 9.2 \scriptsize (+5.4) 
\\
\cmidrule(lr){2-10}

&MLA$\dagger$ \citeyearpar{zhang2022MLA}
&35.8 &21.1
&18.3 &15.2
&30.7 &21.8
&20.4 &18.4
\\

&\quad \bf with TEIR
&\bf 45.9 \scriptsize (+10.1)
&\bf 11.4 \scriptsize (+9.7)
&\bf 25.2 \scriptsize (+6.9)
&\bf 8.7 \scriptsize (+6.5)
&\bf 41.3 \scriptsize (+10.6)
&\bf 12.2 \scriptsize (+9.6)
&\bf 29.2 \scriptsize (+8.8)
&\bf 10.5 \scriptsize (+7.9)
\\
\cmidrule(lr){2-10}

&P-Tuning$\dagger$ \citeyearpar{liu2022P-Tuning}
&29.1 &24.7
&14.5 &16.5
&23.8 &23.8
&15.4 &19.3
\\
&\quad \bf with TEIR
&\bf 41.0 \scriptsize (+11.9)
&\bf 13.1 \scriptsize (+11.6)
&\bf 21.9 \scriptsize (+7.4)
&\bf 9.5 \scriptsize (+7.0)
&\bf 35.0 \scriptsize (+11.2)
&\bf 13.3 \scriptsize (+10.5)
&\bf 24.5 \scriptsize (+9.1)
&\bf 11.0 \scriptsize (+8.3)
\\
\cmidrule(lr){2-10}

&LoRA$\dagger$ \citeyearpar{hu2022LoRA}
&31.3 &23.1
&16.2 &16.0
&27.8 &22.9
&18.8 &18.8

\\
&\quad \bf with TEIR
&\bf 41.6 \scriptsize (+10.3)
&\bf 13.3 \scriptsize (+9.8)
&\bf 22.8 \scriptsize (+6.6)
&\bf 9.8 \scriptsize (+6.2)
&\bf 38.4 \scriptsize (+10.6)
&\bf 13.7 \scriptsize (+9.2)
&\bf 27.2 \scriptsize (+8.4)
&\bf 11.6 \scriptsize (+7.2)
\\
\cmidrule(lr){2-10}

&DualPrompt \citeyearpar{wang2022DualPrompt}  
&27.4 &24.6 &13.9 &16.0
&24.1 &24.3 &15.8 &18.7
\\
&\quad \bf with TEIR
&\bf 38.1 \scriptsize (+10.7)
&\bf 14.3 \scriptsize (+10.3)
&\bf 20.4 \scriptsize (+6.5)
&\bf 10.0 \scriptsize (+6.0)
&\bf 35.1 \scriptsize (+11.0)
&\bf 14.6 \scriptsize (+9.7)
&\bf 24.3 \scriptsize (+8.5)
&\bf 11.5 \scriptsize (+7.2)
\\
\cmidrule(lr){2-10}

&CodaPrompt \citeyearpar{smith2023CodaPrompt}  
&24.8 &27.0 &12.5 &17.2
&20.9 &26.1 &13.4 &20.1
\\
&\quad \bf with TEIR
&\bf 41.0 \scriptsize (+16.2)
&\bf 10.2 \scriptsize (+16.8)
&\bf 22.2 \scriptsize (+9.7)
&\bf 7.3 \scriptsize (+9.9)
&\bf 36.6 \scriptsize (+15.7)
&\bf 9.7 \scriptsize (+16.4)
&\bf 25.4 \scriptsize (+12.0)
&\bf 7.9 \scriptsize (+12.2)
\\

\bottomrule
\end{tabular}
\caption{Retrieval performance on MSCOCO$_{36}$ and XM3600 under the task order (2) presented in Table~\ref{tab:order}. $\dagger$: Task identity is needed during inference. All results are re-produced by ourselves except that of PaLI. Note that PaLI is not optimized for image-text retrieval, but we draw its results from \cite{chen2023PaLI} for completeness. The numbers in brackets indicate the absolute improvements brought by our approach.}
\label{tab:order2_full}
\end{table*}
%%%%%%%%%%%%%%%%

%%%%%%%%%%%%%%%%
\begin{table}[tbp]
\centering
% \tabfootnotesize

\setlength{\tabcolsep}{1.5pt}
% \fontsize{8}{9}\selectfont
\begin{tabular}{@{}llclc@{}}
\toprule 

Setting &Model &Time (h) &Model &Time (h)
\\
\midrule

Joint Lea. &CLL-CLIP$^*$ &13.2 &CLL-CLIP$^\dagger$ &$\approx225$\\ 
\midrule

\multirow{9}{*}{\begin{tabular}[c]{@{}l@{}}Continual\\Learning\end{tabular}}

&(1) CLL-CLIP &11.1 & (1) + TEIR &11.3 \\
&(2) oEWC &14.7 & (2) + TEIR &15.1 \\
&(3) ER &17.6 & (3) + TEIR &18.0 \\
&(4) DER &18.5 & (4) + TEIR &18.8 \\
&(5) MLA &13.5 & (5) + TEIR &13.7 \\
&(6) P-Tuning &13.7 & (6) + TEIR &14.0 \\
&(7) LoRA &12.8 & (7) + TEIR &13.2 \\
&(8) DualPrompt &14.7 & (8) + TEIR &14.9 \\
&(9) CodaPrompt &15.8 & (9) + TEIR &16.1 \\

\bottomrule
\end{tabular}
\caption{Training time of different models measured by a single NVIDIA V100 card. CLL-CLIP$^*$: trained on all 36 tasks (1-time joint learning). CLL-CLIP$^\dagger$: trained on all seen tasks every time a new task arrives (36-times joint learning). 
We can observe that (1) our TEIR only incurs negligible costs on training time; (2) continual learning models are cost-effective compared with the joint-learning CLL-CLIP$^\dagger$ model when dealing with a non-stationary data stream.
}
\label{tab:training_time}
\end{table}
%%%%%%%%%%%%%%%%

%%%%%%%%%%%%%%%%
\begin{figure*}[tbp]
\centering
\includegraphics[width=1\linewidth]{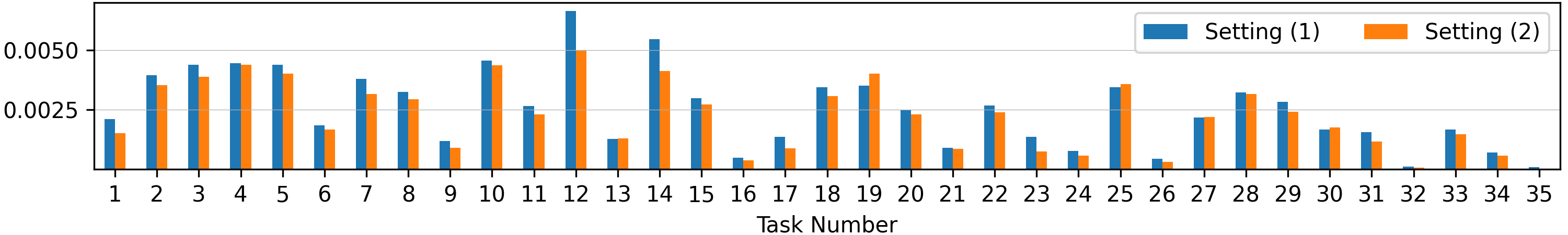}
\caption{KL-divergence between attention distributions of ${\rm Model}_T$ and ${\rm Model}_t$ when processing the training data of the $t$-th task on MSCOCO$_{36}$ ($T=36, t \in [1, T-1]$). ${\rm Model}_t$ denotes a model that has learned $t$ tasks. Lower values indicate that the model's attention pattern is more stable and consistent during continual language learning. Settings (1,2) are from Table~\ref{tab:ablation_TEIR}. We can observe that proper initialization (i.e., setting (2)) can enhance the stability and consistency of the model's attention pattern during CLL.
}
\label{fig:ablation_id}
\end{figure*}
%%%%%%%%%%%%%%%%

\end{document}